\documentclass{article}
\usepackage{amssymb}
\usepackage{amsmath}
\usepackage{mathrsfs}
\usepackage{booktabs}  
\usepackage{adjustbox}
\usepackage{multirow}
\usepackage{makecell}
\usepackage[dvipsnames]{xcolor}

\usepackage[preprint]{corl_2026} 

\title{\textsc{Foci Policy}: Focus on Object-Centric Interactions for Relational Manipulation Policies}

\author{
Ze Fu$^{1,3}$ \quad
Pinhao Song$^{1,3}$ \quad
Yutong Hu$^{1,3}$ \quad
Renaud Detry$^{1,2,3}$ \\
\\
$^1$KU Leuven, Dept. Mechanical Engineering, Research unit Robotics, Automation and Mechatronics \\
$^2$KU Leuven, Dept. Electrical Engineering, Research unit Processing Speech and Images \\
$^3$Flanders Make@KU Leuven \\
\\
\texttt{\{ze.fu,pinhao.song,yutong.hu,renaud.detry\}@kuleuven.be}
}

\begin{document}
\maketitle

\begin{abstract}
Object-centric manipulation policies improve generalization by modeling object motion instead of directly predicting robot actions. However, existing methods are often limited by representations which are either too simplistic to capture interaction dynamics or too dense to learn efficiently. We observe that many rigid relational manipulation tasks are governed by short interaction phases where the relative motion between task-relevant objects is tightly constrained. Based on this observation, we propose \textsc{Foci Policy}, an interaction-centric framework that achieves a two-fold abstraction: (1) temporally, by automatically extracting compact interaction segments from demonstrations;(2) spatially, by representing skills as relative $SE(3)$ motion between task-relevant objects, yielding invariance to scene configurations and robot embodiment. Experiments on RLBench, COLOSSEUM, and real-world tasks show that \textsc{Foci Policy} achieves strong performance with substantially less training data than prior object-centric and action-centric policies. These results suggest that modeling object-object interactions provides a simple and efficient inductive bias for rigid relational manipulation. Project page: \href{https://fitz0401.github.io/foci-page/}{fitz0401.github.io/foci-page/}.
\end{abstract}


\begin{figure}[htb]
    \centering
    \includegraphics[width=1.0\linewidth]{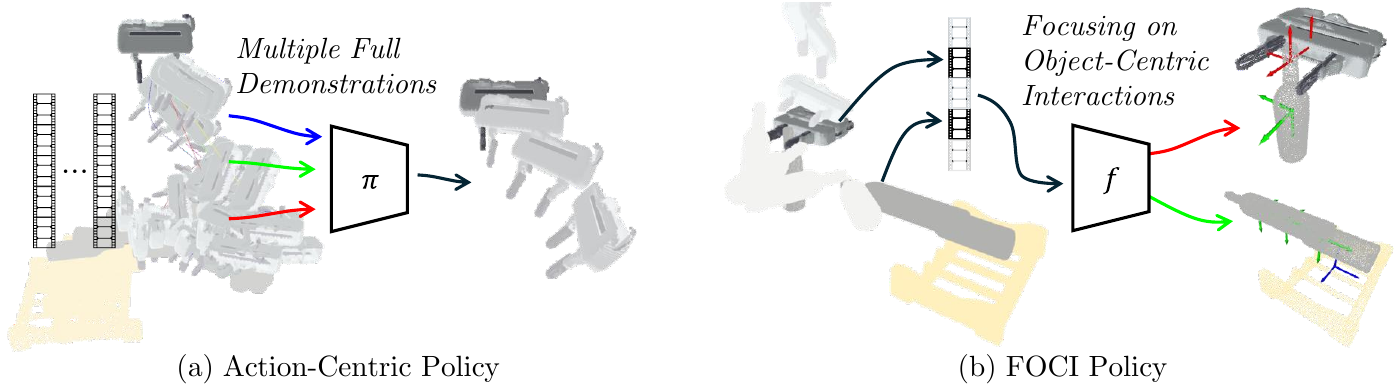}
    \caption{Comparison between action-centric policies and \textsc{Foci Policy}. (a) Action-centric policies learn observation-to-action mappings and require many demonstrations to cover the high-dimensional robot state space. (b) \textsc{Foci Policy} learns only task-relevant interaction segments by predicting relative motion between objects.}
    \label{fig:intro}
\end{figure}


\section{Introduction}
    Humans acquire manipulation skills with remarkable efficiency. From a single demonstration, one can infer how objects should move relative to each other and reproduce the behavior with different embodiments~\cite{peeters2009representation, orban2014neural}. In contrast, most robotic manipulation policies learn a direct mapping from observations to actions~\cite{shridhar2023perceiver, goyal2023rvt, ke20243d, lin2024data}, training a form of ``muscle memory'' without explicitly modeling object--object interactions (OOIs) (Fig.~\ref{fig:intro}~(a)). Such action-centric formulations typically require large numbers of demonstrations across diverse object poses and scene configurations, even for simple pick-and-place tasks.

    Recent works therefore explore object-centric policy learning~\cite{simeonov2023shelving, huang2024imagination, kamil2025sr, xu2024flow, yin2025object, vosylius2023few, yang2026lilo, li2025controlvla}, where robot actions are generated from intermediate object-level representations such as goal poses~\cite{simeonov2023shelving, huang2024imagination, kamil2025sr} or motion flows~\cite{xu2024flow, yin2025object}. While these approaches improve data efficiency and cross-embodiment generalization, existing representations often either oversimplify interactions or model them at unnecessarily high complexity. Static goal poses~\cite{simeonov2023shelving, huang2024imagination, kamil2025sr} fail to capture essential interaction dynamics, whereas dense motion flows~\cite{xu2024flow, yin2025object} are difficult to learn and prone to drift. Our key insight is that many relational manipulation tasks are governed primarily by short intervals of tightly constrained OOIs. For example, when inserting a toothbrush into a cup, the precise relative motion near contact is critical, whereas the preceding transport motion is comparatively unconstrained. This motivates modeling \textit{interaction segments}: compact intervals of constrained relative motion that capture the functional structure of manipulation while discarding task-irrelevant variability.

    Motivated by this observation, we propose \textsc{Foci Policy}, an interaction-centric framework that automatically extracts interaction segments from demonstrations and represents them as relative SE(3) motion between task-relevant objects (Fig.~\ref{fig:intro}~(b)). Conditioned on object point clouds and language instructions, the policy predicts these interaction segments, from which executable robot actions are recovered. Our method performs strongly on relational manipulation tasks in RLBench~\cite{james2020rlbench}, substantially outperforming prior object-centric~\cite{huang2024imagination, hsu2025spot} and action-centric baselines~\cite{ke20243d, goyal2023rvt} despite using only a single demonstration and a single camera. Under severe visual perturbations in COLOSSEUM~\cite{pumacay2024colosseum}, \textsc{Foci Policy} further demonstrates robustness comparable to recent VLA models~\cite{li2025bridgevla, liu2026activevla} trained with 100 demonstrations per task. More broadly, the method achieves 43.7\% average success on RLBench-18 tasks~\cite{shridhar2023perceiver} in the one-demonstration setting.

    
    Our contributions are threefold: (1) \textbf{Interaction as a Compact Abstraction:} We show that task-relevant OOIs provide a compact abstraction for relational manipulation, and propose an automatic pipeline to extract interaction segments from demonstrations. (2) \textbf{Interaction-Centric Policy Architecture:} We design a two-stage policy which models manipulation as relative $SE(3)$ motion between task-relevant objects, yielding invariance to robot embodiment and scene configurations. (3) \textbf{Data-efficient generalization:} The resulting method, \textsc{Foci Policy}, demonstrates strong performance with a single demonstration in both simulation and real-world, suggesting that effective relational manipulation does not necessarily require large-scale training data.


\section{Related Work}
\label{sec:related_work}
    \textbf{Relational Manipulation Policies.} Relational manipulation depends fundamentally on geometric and functional relationships between objects. Standard imitation learning approaches~\cite{shridhar2023perceiver, goyal2023rvt, ke20243d, lin2024data} directly map observations to robot actions, limiting generalization to novel object configurations. Object-centric representations, however, have long been explored in robotics through object-centered control formulations~\cite{khatib2003unified, bruyninckx1996specification} and task-parameterized imitation learning on object frames~\cite{alissandrakis2005approach, calinon2007learning}. Recent advances in vision foundation models~\cite{oquab2023dinov2,ren2024grounded,ravi2024sam,wen2024foundationpose} make such representations practical for real-world manipulation. Existing object-centric policies typically predict either static object goal poses~\cite{simeonov2023shelving, huang2024imagination, kamil2025sr} or dense motion flows~\cite{xu2024flow, yin2025object}, to improve sample efficiency and task generalization. \textsc{Foci Policy} follows this paradigm, but represents manipulation as compact interaction segments between task-relevant objects.
    

    \textbf{Few-Demonstration Imitation Learning.} Learning manipulation policies from only a few demonstrations has received increasing attention due to the high cost of robot data collection~\cite{kamil2025sr}. Prior works improve few-shot adaptation through transferable priors~\cite{pan2023tax, qin2024anyokp, ryu2024diffusion, tang2025functo}, large-scale synthetic pretraining~\cite{vosylius2023few, vosylius2024instant}, or data augmentation~\cite{xue2025demogen}. However, these approaches often depend on domain alignment or augmentation quality. In contrast, \textsc{Foci Policy} achieves data-efficient generalization by exploiting the invariance of object-object interactions across scene configurations.

    \textbf{Temporal Abstraction for OOIs.} Prior works reduce policy learning complexity through temporal abstraction over demonstrations~\cite{shridhar2023perceiver, ke20243d, zhu2024orion, wang2025vlmseerobotdo, ren2025motiontrack}, typically by extracting sparse keyframes from end-effector motion or contact heuristics. However, these methods remain action-centric, and their generalization is still tied to observation-action pairs seen during training. Recent works explore object-centric interaction abstractions~\cite{qi2025twobytwo,Unmesh2024interact}, but their integration into closed-loop manipulation remains limited. Closely related to our work is Imagination Policy~\cite{huang2024imagination}, which predicts target object configurations through dense point-cloud alignment generation. In contrast, \textsc{Foci Policy} performs temporal abstraction by focusing on object-centric interactions, yielding substantially lower cross-demonstration trajectory variance than full trajectory imitation (Appendix~\ref{sec:segment}).

    \textbf{Task and Motion Planning.} Task and motion planning (TAMP) jointly reasons over task structure and motion feasibility to compose manipulation skills, while recent works further integrate learned perception or reusable skills~\cite{liu2024learning, shen2026tiptop, athalye2026pixels}. These methods primarily address how skills or manipulation primitives are selected and composed. In contrast, \textsc{Foci Policy} addresses a complementary problem: learning the continuous object-relative interaction trajectory within a skill directly from demonstrations. This distinction is particularly relevant for interactions such as insertion or articulated-object manipulation that are not naturally expressed by fixed pick-and-place primitives.


\section{Method}
\label{sec:method}
    The core idea of \textsc{Foci Policy} is to distill manipulation skills into its minimal abstraction, which is achieved along two key dimensions: 

    \begin{figure}[tb]
        \centering
        \includegraphics[width=1.0\linewidth]{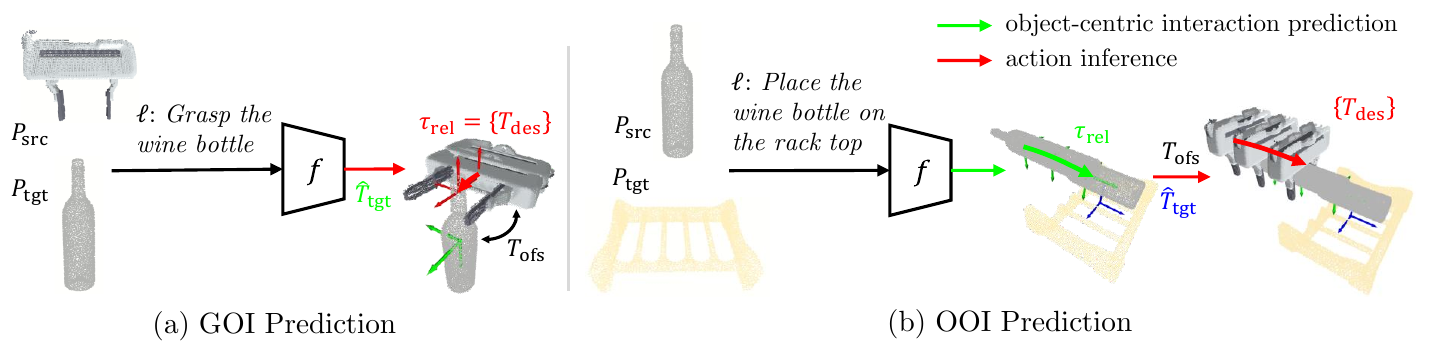}
        \caption{Overview of \textsc{Foci Policy}. Given point clouds of task-relevant objects and a language instruction, the policy first predicts an object-centric relative motion. Then, the desired end-effector trajectory is recovered by (a) \textit{GOI tasks:} composing the predicted relative poses with the estimated target object pose. (b) \textit{OOI tasks:}  composing the predicted interaction with the grasp offset.}
        \label{fig:method}
    \end{figure}
    
    (i) \textbf{Temporal Abstraction via Interaction Segments.} Standard manipulation policies typically predict over an entire demonstration episode $[0, H]$, which includes high-variance transport phases due to arbitrary initial states. We instead isolate the interaction segment $\tau_{\textnormal{rel}}$ within the interval $[t_\textnormal{s}, t_\textnormal{g}] \subset [0, H]$, where object relations are most tightly constrained. This design removes task-irrelevant variability and enforces the learning objective.
    
    (ii) \textbf{Spatial Invariance via Object-Centric Formulation.} To ensure the learned skills are independent of robot embodiment and scene configurations, we represent $\tau_{\textnormal{rel}}$ as the relative motion between task-relevant objects. This formulation reduces the state space to its minimal sufficient abstraction.
    
    Specifically, we define a \emph{source} object as the entity being manipulated (e.g., a grasped tool) and a \emph{target} object as the reference entity (e.g., a container). Let $T_{\textnormal{src}}^{(k)}, T_{\textnormal{tgt}}^{(k)} \in \mathrm{SE}(3)$ denote their world-frame poses. Interaction segments captures the motion of the source object in the target frame:
    \begin{equation}
        \tau_{\textnormal{rel}} = \{T_{\textnormal{rel}}^{(k)}\}_{k=1}^{n} = \{ (T_{\textnormal{tgt}}^{(k)})^{-1} T_{\textnormal{src}}^{(k)} \}_{k=1}^{n}
    \end{equation}
    where $k=1,\dots,n$ indexes discrete waypoints within $[t_s, t_g]$. Gripper-object interactions (GOIs) are treated as a special case where the gripper serves as the source object. By predicting these interaction segments, we can recover robot actions through a deterministic inference pipeline. Given the estimated target pose $\hat{T}_{\textnormal{tgt}}$, we compute the desired end-effector trajectory as
    \begin{equation}
        T_{\textnormal{des}}^{(k)} = T_{\textnormal{ofs}} \, T_{\textnormal{rel}}^{(k)} \, \hat{T}_{\textnormal{tgt}}
    \end{equation}
    Here, $T_{\textnormal{ofs}}$ aligns the source object frame with the end-effector frame. For GOIs (e.g., grasping or pushing), the gripper serves as the source and $T_{\textnormal{ofs}}$ is the identity. For OOIs (e.g., placing or inserting), assuming a rigid grasp, $T_{\textnormal{ofs}}$ corresponds to the transform between the gripper and the grasped object, as illustrated in Fig.~\ref{fig:method}. The final robot action is executed by planning a collision-free path to the initial waypoint $T_{\textnormal{des}}^{(1)}$ and sequentially tracking $\{T_{\textnormal{des}}^{(k)}\}_{k=2}^{n}$.

\subsection{Problem Statement.}
Given expert demonstrations $\mathcal{D}$, we first identify interaction segments $\tau_{\textnormal{rel}}$ (Sec.~\ref{sec: interaction_segment}). Based on that, we learn an interaction-centric policy (Sec.~\ref{sec:model_architecture}). Since multiple valid $\tau_{\textnormal{rel}}$ may exist for a task (e.g., different feasible insertion directions), we model the policy as a conditional distribution. Formally, let $P_{\textnormal{src}}$ and $P_{\textnormal{tgt}}$ denote the point clouds of the source and target objects, and let $\ell$ denote a language instruction. Our objective is to learn a conditional policy $p(\tau_{\textnormal{rel}} \mid P_{\textnormal{src}}, P_{\textnormal{tgt}}, \ell)$. At inference time, executable robot actions are obtained by instantiating $\tau_{\textnormal{rel}}$ in the world frame.

\subsection{Identification of Interaction Segments}
\label{sec: interaction_segment}
\begin{figure}[tb]
        \centering
        \includegraphics[width=0.9\linewidth]{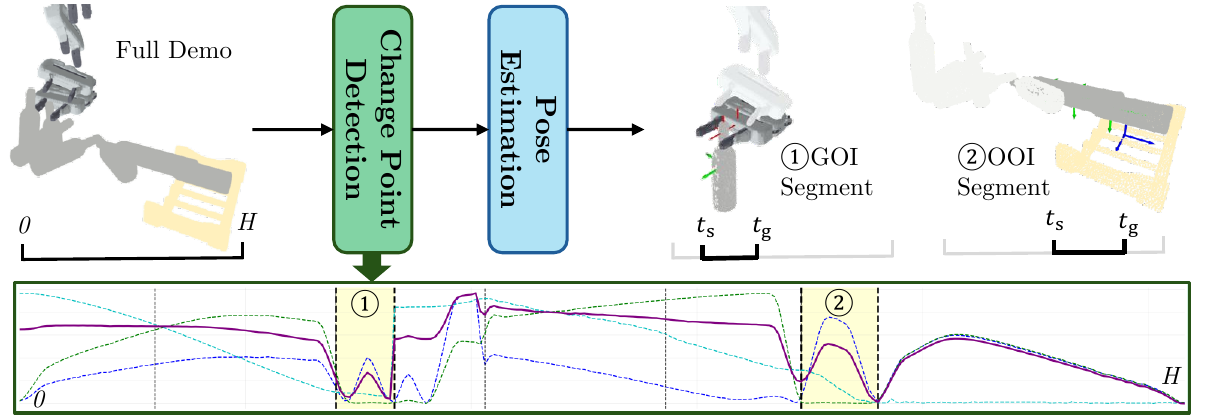}
        \caption{Data processing pipeline. First, change-point detection is applied to each raw demonstration to identify GOI or OOI segments spanning from $t_\textnormal{s}$ to $t_\textnormal{g}$ within the full horizon $[0, T]$. Second, point clouds of task-relevant objects are extracted for each segment, and their poses are estimated and transformed into a relative reference frame. The resulting training sample is represented as $(P_{\textnormal{src}}, P_{\textnormal{tgt}}, \ell, \tau_{\textnormal{rel}})$. \textbf{Bottom:} Example signal from a \textit{Stack-Wine} demonstration. We combine gripper linear velocity (\textcolor{blue}{blue}), angular velocity (\textcolor{Green}{green}), and source--target distance (\textcolor{Cyan}{cyan}) into a unified signal (\textcolor{Purple}{purple}). Detected change points are shown as black dashed lines, and the resulting interaction interval is highlighted in yellow.}
        \label{fig:data}
    \end{figure}

    A key component of \textsc{Foci Policy} is the automatic identification of \emph{interaction segments} (see Fig.~\ref{fig:data}). Prior work~\cite{xue2025demogen, kamil2025sr} and our empirical analysis (Appendix~\ref{sec:segment}) show that relational manipulation demonstrations exhibit a consistent structure: an unconstrained \emph{transport phase} followed by a constrained \emph{interaction phase}. This transition is reflected in changes in kinematic and spatial signals, providing a natural boundary for segment extraction.

    Specifically, we construct a multivariate time series $\tilde{S}(t) = [\tilde{v}(t), \tilde{\omega}(t), \tilde{d}(t)]$, where $v(t)$ and $\omega(t)$ denote the gripper's linear and angular speeds, and $d(t)$ measures the distance between the source and target objects. All signals are normalized over the demonstration horizon.

    We formulate interaction segment identification as a change-point detection (CPD) problem on $\tilde{S}(t)$. For each subtask in $\mathcal{D}$, the goal time $t_\textnormal{g}$ is obtained from gripper state changes or the end of the demonstration, and the start time $t_\textnormal{s}$ is defined as the last change point before $t_\textnormal{g}$. We employ the PELT algorithm~\cite{truong2020selective} to solve this CPD problem, which identifies the optimal set of change points $\mathcal{T} = \{\tau_i\}$ by solving:
    \begin{equation}
        \label{equ:interaction_segment}
        \min_{\mathcal{T}} \sum_{i=0}^{|\mathcal{T}|} C\big(\tilde{S}(\tau_i:\tau_{i+1})\big) + \beta |\mathcal{T}|,
    \end{equation}
    where $C(\cdot)$ is a segment cost measuring within-segment variation, implemented using an RBF kernel to capture changes in the signal, and $\beta$ controls the number of change points. This lightweight approach isolates the functionally critical interaction segments without requiring manual annotation.

\subsection{Object-Centric Interaction Prediction}
    \label{sec:model_architecture}
    \textbf{Data Preparation.} To support interaction-centric learning, from each demonstration, we estimate object poses $T_{\textnormal{src}}, T_{\textnormal{tgt}}$ from observations and express the motion of the source object in the target frame as $T_{\textnormal{rel}} = T_{\textnormal{tgt}}^{-1} T_{\textnormal{src}}$. Both object point clouds are transformed into their canonical frames, yielding an object-centric representation that is invariant to scene configuration. We then extract interaction segments and uniformly sample $n$ waypoints within $[t_\textnormal{s}, t_\textnormal{g}]$ to form $\tau_{\textnormal{rel}}$. The final training samples are constructed as $(P_{\textnormal{src}}, P_{\textnormal{tgt}}, \ell, \tau_{\textnormal{rel}})$.
    
\begin{figure}[tb]
        \centering
        \includegraphics[width=0.9\linewidth]{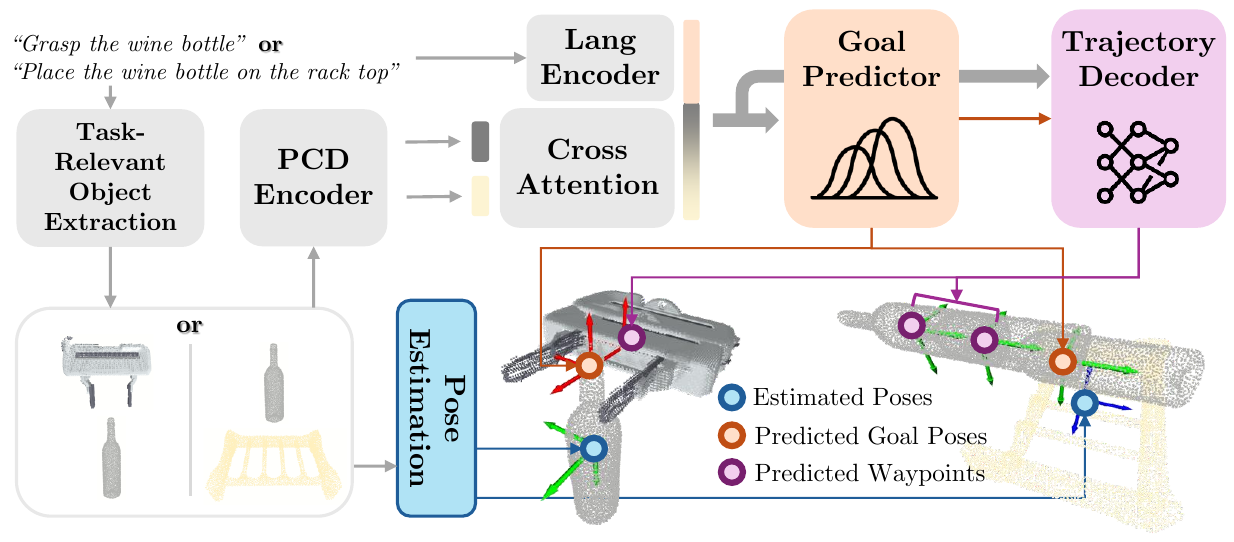}
        \caption{Model architecture and inference pipeline. Given a language instruction, our method extracts task-relevant objects and encodes their point clouds. The resulting features are fused via cross-attention and combined with language embeddings. A goal predictor parameterized as GMM estimates a feasible interaction goal (\textcolor{Orange}{orange}), which conditions a trajectory decoder to generate intermediate waypoints (\textcolor{Purple}{purple}). The bottom visualizations depict two distinct cases: a \textbf{GOI}(left) showing the gripper approaching a bottle, and an \textbf{OOI}(right) showing a bottle being placed on a rack. During inference, the predicted trajectory is instantiated in the world frame using the estimated object pose (\textcolor{Blue}{blue}) to produce executable robot actions.}
        \label{fig:model_arch}
    \end{figure}

    \textbf{Model Architecture.} Our model is designed to capture the multi-modal nature of interactions while maintaining geometric precision. We learn a conditional model $p(\tau_{\textnormal{rel}} \mid P_{\textnormal{src}}, P_{\textnormal{tgt}}, \ell)$ that maps object-centric observations and language instructions to interaction segments (Fig.~\ref{fig:model_arch}).

    Given $P_{\textnormal{src}}$ and $P_{\textnormal{tgt}}$, we encode local geometric features using point-cloud encoders with identical architectures. During training, we randomly drop a subset of point-cloud tokens before feature fusion, encouraging robustness to partial observations. Cross-attention is then applied between source and target features to capture interaction-relevant geometry. The resulting tokens are concatenated with the language embedding and fused via a Transformer encoder layer to obtain the context representation $H$. Conditioned on this context, we predict the interaction segment in two stages:
    \begin{equation}
    p(\tau_{\textnormal{rel}} \mid H)
    = \underbrace{p_{\text{goal}}\big(T^{(n)}_{\textnormal{rel}} \mid H\big)}_{\text{Stage 1: Goal Prediction}}
    \cdot
    \underbrace{p_{\text{traj}}\big(\{T^{(k)}_{\textnormal{rel}}\}_{k=1}^{n-1} \mid H, T^{(n)}_{\textnormal{rel}}\big)}_{\text{Stage 2: Trajectory Generation}}.
    \end{equation}
    Stage 1 predicts a multi-modal goal pose $T^{(n)}_{\textnormal{rel}}$ through GMM. The architecture can represent multiple valid goal modes when such modes are present in the demonstrations. In Stage 2, a trajectory decoder generates intermediate poses conditioned on the predicted goal. The model is trained to match the ground-truth interaction segment. We optimize a combination of (i) a goal likelihood loss for terminal pose prediction and (ii) a trajectory reconstruction loss for intermediate waypoints. We provide training details in Appendix.~\ref{app:implementation_details}.

    Restricting learning to $[t_\textnormal{s}, t_\textnormal{g}]$ in object-centric frames reduces observation variance and abstracts away task-irrelevant motion, leading to improved accuracy of interaction modeling.
    

\section{Experiments}
\label{sec:experiments}
    We evaluate \textsc{Foci Policy} to answer four research questions: \textbf{(RQ1)} Can interaction-centric representations enable data-efficient learning on relational manipulation tasks with clear OOIs? \textbf{(RQ2)} Does such a formulation improve robustness under severe visual distribution shifts? \textbf{(RQ3)} How well does the formulation generalize beyond tasks with clearly structured OOIs? \textbf{(RQ4)} Can the formulation remain effective under real-world perceptual uncertainty and execution noise?

    To answer these questions, we use three complementary simulation benchmarks: We address RQ1 with experiments on 8 tasks selected from RLBench~\cite{james2020rlbench} for their OOI nature. We address RQ2 with COLOSSEUM~\cite{pumacay2024colosseum} (robustness under diverse visual perturbations). We address RQ3 with the RLBench-18 tasks~\cite{shridhar2023perceiver} (generalization across a broader range of manipulation skills). Finally, we validate our work on five real-robot tasks under increasing levels of pose variation (RQ4).


    \subsection{Simulation Experiment}
    \textbf{Implementation Details.} We train separate multi-task models ($f_{\text{GOI}}$ and $f_{\text{OOI}}$) with identical architectures. Object point clouds are encoded using PointNet++~\cite{qi2017pointnet++}, and language instructions via a pretrained CLIP-ViT32~\cite{radford2021learning}. For each task, we provide one-time semantic specifications consisting of the source/target object names and brief phase descriptions. Interaction boundaries are then detected automatically via change-point detection using a single global penalty ($\beta=0.6$; Appendix~\ref{sec:segment}) without task-specific tuning. For RLBench, object poses are estimated using FoundationPose~\cite{wen2024foundationpose} initialized from object meshes and ground-truth masks. For COLOSSEUM,masks are extracted from RGB images using GroundedSAM~\cite{ren2024grounded} and then provided to FoundationPose together with the object meshes. Our two-stage architecture predicts a GMM-parameterized goal distribution followed by intermediate relative poses. All models are trained using AdamW on a single NVIDIA RTX 2080 Ti GPU. All reported experiments use open-loop execution; pose tracking and replanning are not enabled. Additional details are provided in Appendix~\ref{app:implementation_details}.

    \textbf{Test on Relational Manipulation Benchmark.} To answer RQ1, we evaluate on 8 RLBench tasks characterized by clear OOIs that are suitable for interaction segment extraction, including insertion, articulated manipulation, and placement under clutter. We consider two variants of \textsc{Foci Policy}: one trained on a single demonstration, and another trained on five demonstrations. All baselines~\cite{huang2024imagination, hsu2025spot, ke20243d, goyal2023rvt} use 5 demonstrations. Results are summarized in Table~\ref{tab:simu_results}. \textsc{Foci Policy} outperforms both object-centric~\cite{huang2024imagination, hsu2025spot} and action-centric~\cite{ke20243d, goyal2023rvt} baselines while requiring only a single camera and less than 0.5 hours of training, particularly on precision-critical tasks such as \textit{Put-Roll} and \textit{Put-Umbrella}. These results confirm that interaction-centric representations provide an effective inductive bias for relational manipulation when OOIs are reliably extractable.

    \begin{figure}[tb]
        \centering
        \includegraphics[width=1.0\linewidth]{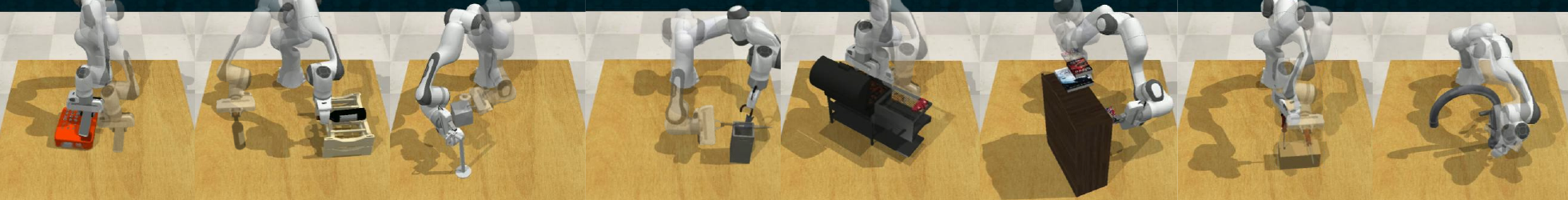}
        \caption{Eight relational manipulation tasks in RLBench. From left to right and top to bottom: \textit{Phone-on-Base}, \textit{Stack-Wine}, \textit{Put-Roll}, \textit{Put-Umbrella}, \textit{Meat-on-Grill}, \textit{Put-Books}, \textit{Screw-Nail}, and \textit{Turn-Tap}. Each image visualizes the OOI phase, with a transparent overlay showing the GOI phase.}
        \label{fig:simu_exp}
    \end{figure}
    
    \begin{table*}[tb]
    \centering
    \caption{Performance on 8 relational manipulation tasks in RLBench. Success rate (\%) is evaluated over 25 unseen tests and averaged across 3 runs. When task-relevant OOIs are well-defined and perceptually grounded, \textsc{Foci Policy} achieves strong performance with only 1 demonstration and a single RGB-D camera, outperforming prior baselines trained with more demonstrations and multi-view observations. Training time is measured on a single NVIDIA RTX 2080 Ti GPU.}
    \label{tab:simu_results}
    \begin{adjustbox}{max width=\textwidth}
    \small
    \begin{tabular}{l|c|c|cccccccc|c}
        \toprule
        Method & Demo & Cam & phone\_on\_base & stack\_wine & put\_roll & put\_umbrella & meat\_on\_grill & put\_books & screw\_nail & turn\_tap & Time. \\
        \midrule
        \textsc{Foci Policy} (Ours) & 1 & 1 & \underline{96.0 ± 3.3} & \underline{97.3 ± 1.9} & \textbf{45.3 ± 5.0} & \underline{37.7 ± 4.8} & \underline{98.7 ± 1.9} & \textbf{54.7 ± 8.2} & \underline{42.7 ± 1.9} & \underline{96.0 ± 0.0} & \textbf{0.45 h} \\
        \textsc{Foci Policy} (Ours) & 5 & 1 & \textbf{97.3 ± 1.9} & \textbf{100.0 ± 0.0} & \textbf{45.3 ± 5.0} & \textbf{52.0 ± 3.3} & \textbf{100.0 ± 0.0} & \underline{53.3 ± 1.9} & \textbf{53.3 ± 5.0} & \textbf{100.0 ± 0.0} & \underline{0.53 h} \\
        Imagination Policy & 5 & 4 & 73.3 ± 7.5 & 88.0 + 3.3 & \underline{8.7 ± 3.6} & 33.3 ± 5.0 & 92.0 ± 3.3 & 5.3 ± 1.9 & 2.7 ± 1.9 & 0.0 & -\\
        Imagination Policy & 5 & 1 & 57.3 ± 3.8 & 72.0 ± 3.3 & 5.8 ± 2.1 & 9.3 ± 1.9 & 56.0 ± 0.0 & 5.3 ± 5.0 & 4.0 ± 5.7 & 0.0 & 6.58 h\\
        SPOT (w/ GT Pose) & 5 & 1 & 26.7 ± 1.9 & 18.7 ± 1.9 & 0.0 & 0.0 & 86.7 ± 5.0 & 0.0 & 18.7 ± 1.9 & 90.7 ± 3.8 & 2.78 h\\
        3D Diffuser Actor & 5 & 4 & 74.7 ± 5.0 & 93.3 ± 5.0 & 0.0 & 9.3 ± 3.8 & 68.0 ± 3.3 & 32.0 ± 3.3 & 13.3 ± 8.2 & 40.0 ± 3.8 & - \\
        3D Diffuser Actor & 5 & 1 & 62.7 ± 1.9 & 89.3 ± 5.0 & 0.0 & 4.0 ± 3.3 & 50.7 ± 1.9 & 17.3 ± 1.9 & 1.3 ± 1.9 & 30.7 ± 6.5 & 8.97 h\\
        RVT2 & 5 & 4 & 25.3 ± 8.2 & 21.3 ± 9.4 & 0.0 & 0.0 & 4.0 ± 3.3 & 21.3 ± 13.2 & 2.7 ± 1.9 & 53.3 ± 6.8 & - \\
        RVT2 & 5 & 1 & 8.0 ± 3.3 & 17.3 ± 1.9 & 0.0 & 0.0 & 12.0 ± 5.7 & 20.0 ± 14.2 & 1.3 ± 1.9 & 29.3 ± 3.8 & 4.13 h\\
        \bottomrule
    \end{tabular}
    \end{adjustbox}
    \end{table*}

    \textbf{Test on COLOSSEUM.} To answer RQ2, we evaluate our method under six relational manipulation tasks with clear OOIs in COLOSSEUM~\cite{pumacay2024colosseum}, under 12 categories of unseen visual perturbations including texture, lighting, distractor, and camera pose changes. We compare against state-of-the-art 3D VLA models~\cite{li2025bridgevla, liu2026activevla}. Results are shown in Table~\ref{tab:colosseum_results}. Although VLA models achieve higher nominal performance (96.9\% vs. 58.9\%), \textsc{Foci Policy} achieves over 60\% of their absolute performance using only 1\% of the data. Notably, under the most severe perturbation setting (All Perturbations, detailed results in Appendix~\ref{app:results_colosseum}), FOCI POLICY outperforms both VLA models (25.3\% vs. 20.1\%/21.4\%), supporting the robustness of $SE(3)$ object-centric representations under severe visual distribution shifts.

    \begin{table*}[tb]
    \centering
    
    \begin{minipage}[t]{0.48\textwidth}
    \centering
    \caption{Robustness under distribution shifts on six tasks in COLOSSEUM. While VLA models achieve higher performance in nominal settings, our method shows stronger robustness under severe perturbations despite using less data.}
    \label{tab:colosseum_results}
    \small
    \begin{tabular}{lccc}
        \toprule
        Method & Demo & No Var. & All Pert. \\
        \midrule
        \textsc{Foci Policy} & \textbf{1} & 58.9 & \textbf{25.3 ($\downarrow$ 57.0\%)} \\
        BridgeVLA & 100 & 96.5 & 20.1 ($\downarrow$ 79.2\%) \\
        ActiveVLA & 100 & \textbf{96.9} & 21.4 ($\downarrow$ 77.9\%) \\
        \bottomrule
    \end{tabular}
    \end{minipage}
    \hfill
    \begin{minipage}[t]{0.48\textwidth}
    \centering
    
    \caption{Average success rates on RLBench-18 tasks (1-demo setting), grouped by interaction segment grounding difficulty. ``w/ GT'' reports upper-bound performance with ground-truth poses.}
    \label{tab:simu_results_rlbench18}
    \small
    \begin{tabular}{l |cc|c}
        \toprule
        Method & Easy & Hard & Overall \\
        \midrule
        \textsc{Foci Policy}     & \textbf{79.6} & \textbf{7.8}  & \textbf{43.7} \\
        \quad(w/ GT Pose)        & \textbf{98.2} & \textbf{78.1} & \textbf{88.2} \\
        Imagination Policy       & 35.4          & 2.7           & 19.0 \\
        \bottomrule
    \end{tabular}
    \end{minipage}
    \end{table*}

    \textbf{Test on RLBench-18 Tasks.} To answer RQ3, we benchmark \textsc{Foci Policy} on the RLBench-18 suite~\cite{shridhar2023perceiver}, analyzing performance across tasks stratified by interaction segment grounding difficulty (taxonomy and detailed results in Appendix~\ref{app:rlbench_18_details}). Results are summarized in Table~\ref{tab:simu_results_rlbench18}. On easy-to-ground tasks, \textsc{Foci Policy} achieves 79.6\% success with only a single demonstration, substantially outperforming the baseline~\cite{huang2024imagination} in the low-data regime. On hard-to-ground tasks (e.g., thin objects such as \textit{Insert-Peg}, or heavily occluded scenes such as \textit{Put-Groceries}), raw performance drops to 7.8\%; replacing estimated poses with ground-truth raises this to 78.1\%, confirming that the bottleneck is reliable interaction grounding rather than the formulation itself. Overall, the results suggest that interaction-centric policies remain effective beyond tasks with clearly defined OOIs, while also highlighting the importance of robust interaction grounding under occlusion and ambiguous geometry. Detailed failure analysis is provided in Appendix~\ref{sec:failure}.


    \textbf{Ablation Study.} We further evaluate two key components of \textsc{Foci Policy}: interaction segment modeling and the two-stage model architecture. We compare against variants without segmentation, with fixed-length segments, and a one-stage trajectory predictor without an explicit goal model (details in Appendix~\ref{app:ablation}). Results in Table~\ref{tab:ablation} show that interaction segment modeling is critical. Removing segmentation leads to a substantial performance drop, particularly on precision-sensitive tasks. In addition, the two-stage architecture consistently outperforms the one-stage variant, indicating the benefit of explicitly modeling the terminal goal for relational manipulation.

    \begin{table}[tb]
    \centering
    \caption{Ablation study on interaction segment modeling and architecture design. Results report mean success rates (\%) over five RLBench tasks.}
    \label{tab:ablation}
    \begin{adjustbox}{max width=\textwidth}
    \begin{tabular}{l | ccccc | c}
        \toprule
        Model Variant & phone\_on\_base & stack\_wine & put\_umbrella & put\_roll & put\_books & Avg. \\
        \midrule
        \textbf{\textsc{Foci Policy} (Default)} & \textbf{97.3} & \textbf{100.0} & \textbf{52.0} & 45.3 & \textbf{53.3} & \textbf{69.6} \\
        \midrule
        \quad w/o Interaction Segment & 77.3 & 89.3 & 0.0 & 25.3 & 1.3 & 38.6 \\
        \quad Fixed Length (20 frames) & 93.3 & \textbf{100.0} & 44.0 & 38.7 & 46.7 & 64.5 \\
        \quad Fixed Length (10 frames) & 81.3 & 96.0 & 17.3 & \textbf{46.7} & 8.0 & 49.9 \\
        \quad One-stage Trajectory Prediction & 90.7 & 90.7 & 49.3 & 16.0 & 50.7 & 59.5 \\
        \bottomrule
    \end{tabular}
    \end{adjustbox}
    \end{table}

    \subsection{Real Robot Experiment}
    \textbf{Settings.} To answer RQ4, we deploy \textsc{Foci Policy} on a Franka Emika Panda equipped with a single front-facing RealSense L515 RGB-D camera. Demonstrations are collected via kinesthetic teaching, recording synchronized RGB-D observations and gripper states. Object masks are initialized with GroundedSAM~\cite{ren2024grounded} and tracked with XMem++~\cite{bekuzarov2023xmem}. A single demonstration is provided per task to train a multi-task policy. During inference, robot actions are executed using cuRobo~\cite{sundaralingam2023curobo}.

    \begin{figure}[tb]
        \centering
        \includegraphics[width=1.0\linewidth]{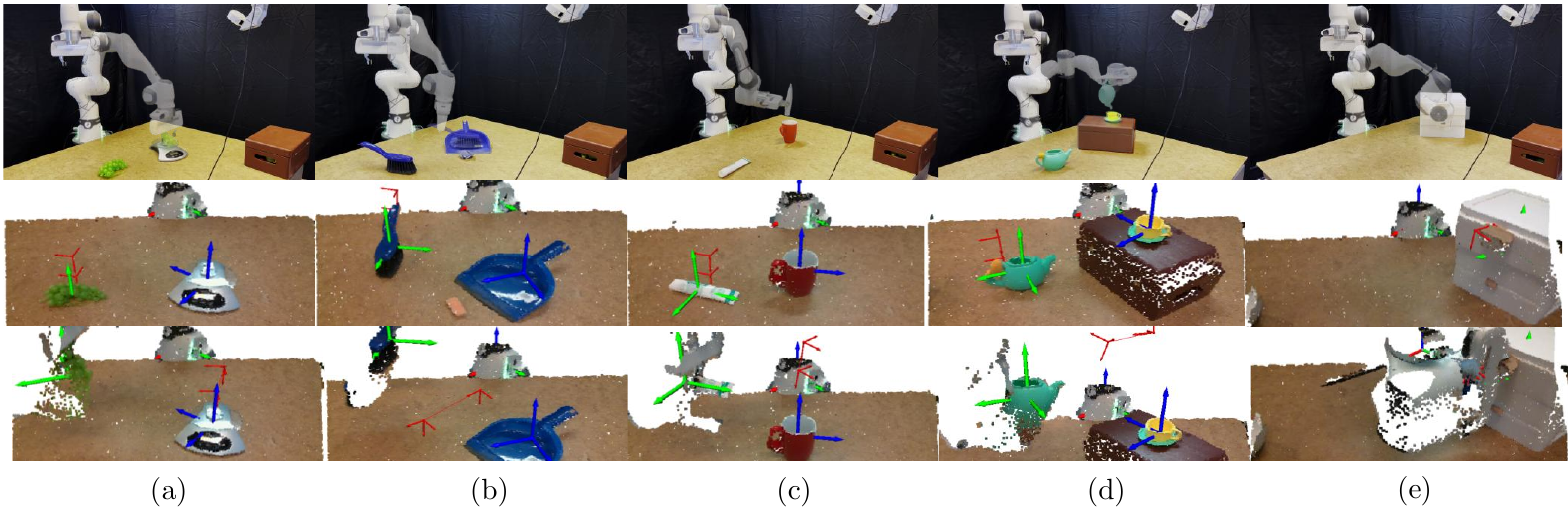}
        \caption{Real-robot tasks. The first row shows the initial and goal states, with the goal state transparently overlaid. The second and third rows visualize the predicted interaction segments in the GOI and OOI frames, respectively. \textcolor{green}{Green} frames denote the source object pose, \textcolor{blue}{blue} frames denote the target object pose, and \textcolor{red}{red} frames denote the predicted robot actions. Tasks from left to right: (a) Scale-Grape, (b) Sweep-Dust, (c) Insert-Tube, (d) Pour-Liquid, and (e) Open-Drawer.}
        \label{fig:real_world}
    \end{figure}
    
    \textbf{Tasks and Baselines.} We evaluate on five tasks spanning tool use (\textit{Sweep-Dust}), precise insertion (\textit{Insert-Tube}), pouring (\textit{Pour-Liquid}), placement (\textit{Scale-Grape}), and articulated manipulation (\textit{Open-Drawer}). Each task is evaluated over 15 trials with increasing difficulty, from small pose perturbations to larger spatial variations. We compare against \textit{Replay Policy}~\cite{huang2025match}, which transfers keyframe gripper poses from the demonstration to novel scenes via Generalized ICP~\cite{segal2009icp}.
    
    \textbf{Results.} Results are reported in Table~\ref{tab:real_results}. \textsc{Foci Policy} achieves over 53\% task success across all tasks with a single demonstration, outperforming the replay baseline on all five tasks. The advantage is most pronounced on precision-critical tasks (\textit{Insert-Tube}, \textit{Pour-Liquid}), where rigid pose transfer degrades under sparse or partially occluded point clouds. These results confirm that the interaction-centric representation transfers effectively to real-world deployment under perceptual uncertainty. We further show in Appendix~\ref{app:real_world_details} that a new skill can be learned and deployed within 10 minutes.

    \begin{table}[tb]
    \centering
    \caption{Success rates on real-world tasks. For each task, we report GOI success, OOI success, and full task success rate. With only one demonstration and a single-camera setup, \textsc{Foci Policy} consistently achieves over 53\% task success across all tasks and variations.}
    \label{tab:real_results}
    \begin{adjustbox}{max width=\textwidth}
    \begin{tabular}{l | c c c c c}
    \toprule
    Method & scale\_grape & sweep\_dust & insert\_tube & pour\_liquid & open\_drawer \\
    \midrule
    \textsc{Foci Policy} (Ours)
    & \makecell{ GOI: 13/15\\ OOI: 12/13\\ \textbf{Task: 12/15 (80\%)}}
    & \makecell{ GOI: 13/15\\ OOI: 11/13\\ \textbf{Task: 11/15 (73\%)}}
    & \makecell{ GOI: 13/15\\ OOI: 9/13\\ \textbf{Task: 9/15 (60\%)}}
    & \makecell{ GOI: 11/15\\ OOI: 8/11\\ \textbf{Task: 8/15 (53\%)}}
    & \makecell{ GOI: 11/15\\ OOI: 9/11\\ \textbf{Task: 9/15 (60\%)}} \\
    \midrule
    Replay Policy
    & \makecell{ GOI: 5/15\\ OOI: 4/5\\ Task: 4/15 (27\%)}
    & \makecell{ GOI: 4/15\\ OOI: 4/4\\ Task: 4/15 (27\%)}
    & \makecell{ GOI: 6/15\\ OOI: 4/6\\ Task: 4/15 (27\%)}
    & \makecell{ GOI: 6/15\\ OOI: 1/6\\ Task: 1/15 (7\%)}
    & \makecell{ GOI: 7/15\\ OOI: 7/7\\ Task: 7/15 (47\%)} \\
    \bottomrule
    \end{tabular}
    \end{adjustbox}
    \end{table}



\section{Conclusion}
\label{sec:conclusion}
    We present \textsc{Foci Policy}, an interaction-centric framework that distills relational manipulation into compact relative $SE(3)$ motion between task-relevant objects. By automatically extracting interaction segments via change-point detection and modeling them through a two-stage architecture, the method achieves strong one-shot generalization across diverse simulation benchmarks and real-world tasks. Beyond data efficiency, the formulation also naturally supports transfer across scene configurations and robot embodiments. Our results suggest that, for rigid relational manipulation, policies can be learned effectively without large-scale demonstrations by explicitly modeling interaction segments.
    
    \textbf{Limitations.} \textit{Formulation scope.} Our framework assumes that task-relevant interaction phases are temporally separable and geometrically identifiable, which is well satisfied in relational manipulation tasks with clear OOIs but less applicable to tasks involving continuous contact (e.g., Push-T~\cite{chi2024dp}). Additionally, learning from a single demonstration limits interaction mode coverage: tasks admitting multiple valid interaction strategies benefit from broader demonstration diversity. \textit{Perceptual reliability.} \textsc{Foci Policy} relies on pretrained object segmentation and pose-estimation modules to construct its object-centric representation. Performance degrades when object geometry is ambiguous or weakly observable, as unreliable canonicalization directly propagates to interaction prediction (see Appendix~\ref{sec:failure} for detailed analysis). Future work may address this through reconstruction-based or zero-shot pose estimation~\cite{lin2024sam6d, zhang2024omni6dpose, wang2024dust3r, wen2023bundlesdf}, reducing reliance on object meshes and improving robustness in challenging perceptual conditions. Extending the framework to deformable objects also remains an important direction for future work~\cite{moghani2026softmimicgen}.
    

\newpage
\section*{Appendix}
\appendix
\section{Additional Experimental Results and Details}
\subsection{Results of Interaction Segments}
\label{sec:segment}
\textbf{Choice of Interaction Segment Parameter.} In Eq.~\ref{equ:interaction_segment}, the penalty factor $\beta$ controls the number of detected change points in the CPD algorithm. Intuitively, a larger $\beta$ produces fewer change points and therefore longer interaction intervals, which better preserve complete interaction dynamics but introduce higher trajectory variance and less compact representations. In contrast, a smaller $\beta$ yields shorter intervals with lower variance, but risks omitting important interaction motions. Selecting $\beta$ therefore requires balancing interaction completeness against trajectory compactness.

To avoid per-task tuning in the few-demonstration regime, we calibrate a single global $\beta$ once using an offline development set. Specifically, we collect 10 additional demonstrations for each RLBench-18 task; these demonstrations are disjoint from the one-demo policy training data and are used only for selecting $\beta$, never for policy training or evaluation. For every trajectory, we uniformly resample 50 waypoints and normalize them along the temporal axis. We then compute the cross-demonstration translation and rotation standard deviation of the gripper trajectory as an approximation of trajectory variance before abstraction. Next, following Sec.~\ref{sec: interaction_segment}, we solve CPD using $\beta \in [0,1]$ with step size $0.1$. For each resulting interaction segment (grasping and manipulation), the gripper trajectory is transformed into the corresponding source or target object frame, and the same variance statistics are recomputed within the extracted interval.

\begin{figure}[htb]
    \centering
    \includegraphics[width=1.0\linewidth]{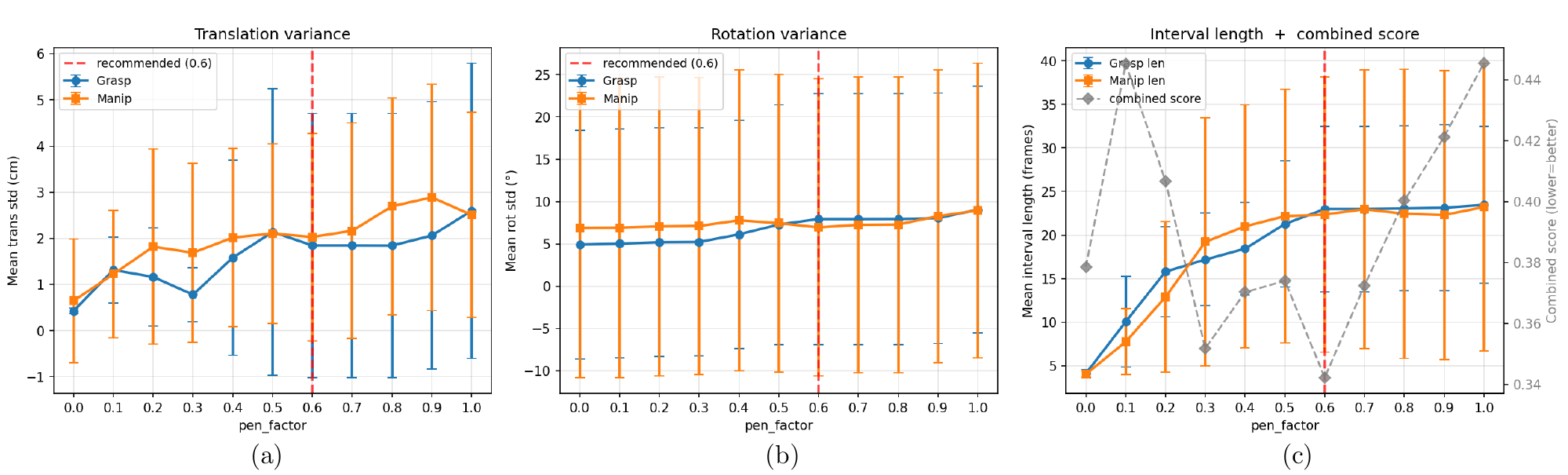}
    \caption{Cross-task selection of the CPD penalty factor $\beta$ on RLBench-18. For each task, 10 demonstrations are collected to estimate cross-demonstration trajectory variance. (a) and (b) plots show the mean translation and rotation variance of extracted interaction segments under different $\beta$ values, respectively. (c) shows the resulting interaction interval length and the normalized combined score balancing variance reduction and segment compactness. Averaged across all 18 tasks, $\beta=0.6$ achieves the best overall trade-off and is used in all experiments.}
    \label{fig:pen_factor_exp}
\end{figure}

To balance compactness and consistency, we define a combined score consisting of translation variance, rotation variance, and normalized interval length. For each task, scores across different $\beta$ values are normalized to $[0,1]$. We then average the normalized scores across all RLBench-18 tasks and select the $\beta$ with the lowest mean score. Results are shown in Fig.~\ref{fig:pen_factor_exp}. Across RLBench-18 tasks, $\beta=0.6$ provides the best trade-off between trajectory variance reduction and interaction interval compactness.

\textbf{Verification of Interaction Segment Detection.} We first verify whether the extracted interaction segments indeed reduce cross-demonstration trajectory variability. For each RLBench-18 task, we collect 30 demonstrations and compute the translation and rotation variance across trajectories. Results are summarized in Table~\ref{tab:traj_std}. Compared with full demonstration trajectories, both GOI and OOI interaction segments exhibit substantially lower variance across nearly all tasks, with average reductions of 78.6\%/90.5\% and 73.4\%/92.3\% in translation/rotation variance, respectively. These results confirm that the proposed interaction segmentation successfully isolates the geometrically constrained phases of manipulation, providing a substantially more compact and consistent learning target than full action trajectories. This observation also explains why the interaction-centric formulation enables strong generalization under extremely limited demonstrations.

\begin{table}[tb]
\centering
\caption{Cross-demonstration trajectory variability on RLBench-18. \textbf{Full} denotes the entire trajectory, while \textbf{GOI} and \textbf{OOI} denote the extracted interaction segments. Translation and rotation standard deviations are computed across 30 demonstrations per task. Interaction segments consistently exhibit substantially lower variance than full trajectories, supporting the interaction-centric formulation.}
\label{tab:traj_std}
    \resizebox{\linewidth}{!}{
    \begin{tabular}{l|cc|cccc|cccc}
    \toprule
    \multirow{2}{*}{Task} 
    & \multicolumn{2}{c|}{Full} 
    & \multicolumn{4}{c|}{GOI} 
    & \multicolumn{4}{c}{OOI} \\
    & \makecell{Trans.\\Std (cm)} & \makecell{Rot.\\Std ($^\circ$)}
    & \makecell{Trans.\\Std (cm)} & \makecell{Rot.\\Std ($^\circ$)}
    & \makecell{Trans.\\$\downarrow$\%} & \makecell{Rot.\\$\downarrow$\%}
    & \makecell{Trans.\\Std (cm)} & \makecell{Rot.\\Std ($^\circ$)}
    & \makecell{Trans.\\$\downarrow$\%} & \makecell{Rot.\\$\downarrow$\%} \\
    \midrule
    close\_jar       & 12.07 & 16.76 & 1.18 & 0.61 & 90.2 & 90.4 & 6.66 & —    & 44.8       & —    \\
    meat\_off\_grill & 16.36 & 61.80 & 2.54 & 2.08 & 84.5 & 96.6 & 6.23 & 0.12 & 61.9       & 99.8 \\
    stack\_wine      & 7.77  & 8.92  & 2.84 & 1.97 & 63.4 & 77.9 & 0.68 & 0.55 & 91.2       & 93.8 \\
    put\_in\_safe    & 9.89  & 7.34  & 0.44 & 0.16 & 95.5 & 97.8 & 7.43 & 1.74 & 24.9       & 76.3 \\
    turn\_tap        & 12.81 & 55.33 & 9.84 & 8.70 & 23.2 & 84.3 & 0.59 & 30.60& 95.4       & 44.7 \\
    put\_groceries   & 19.20 & 42.80 & 4.20 & 2.59 & 78.1 & 93.9 & 6.25 & 4.22 & 67.4       & 90.1 \\
    drag\_stick      & 13.04 & 53.43 & 0.55 & 0.45 & 95.8 & 99.2 & 0.84 & 0.24 & 93.6       & 99.6 \\
    screw\_bulb      & 15.89 & 47.61 & 1.07 & —    & 93.3 & —    & 2.22 & 0.12 & 86.0       & 99.7 \\
    place\_cups      & 12.31 & 42.60 & 0.71 & 0.30 & 94.2 & 99.3 & 0.74 & 0.38 & 94.0       & 99.1 \\
    sweep\_dust      & 14.64 & 58.99 & 0.61 & 0.46 & 95.8 & 99.2 & 4.06 & 0.09 & 72.3       & 99.8 \\
    open\_drawer     & 9.23  & 5.99  & 0.65 & 0.51 & 93.0 & 91.5 & 0.31 & 0.02 & 96.6       & 99.7 \\
    sort\_shape      & 14.54 & 57.05 & 0.54 & 0.36 & 96.3 & 99.4 & 2.05 & 0.63 & 85.9       & 98.9 \\
    insert\_peg      & 15.67 & 54.38 & 0.52 & 0.26 & 96.7 & 99.5 & 3.74 & 4.78 & 76.1       & 91.2 \\
    push\_buttons    & 11.34 & 0.13  & 6.18 & —    & 45.5 & —    & 0.61 & —    & 94.6       & —    \\
    slide\_color     & 6.81  & 43.78 & 5.81 & 30.72& 14.7 & 29.8 & 0.46 & 0.06 & 93.2       & 99.9 \\
    put\_in\_drawer  & 6.23  & 9.65  & 1.80 & 0.47 & 71.1 & 95.1 & 7.60 & 0.05 & $\uparrow$22.0 & 99.5 \\
    stack\_blocks    & 16.72 & 59.59 & 1.40 & 1.96 & 91.6 & 96.7 & 4.20 & 8.45 & 74.9       & 85.8 \\
    stack\_cups      & 14.50 & 61.14 & 1.16 & 1.78 & 92.0 & 97.1 & 1.45 & 0.24 & 90.0       & 99.6 \\
    \midrule
    \textbf{Average} & \textbf{12.72} & \textbf{38.18} 
    & \textbf{2.22} & \textbf{3.38} & \textbf{78.6} & \textbf{90.5}
    & \textbf{3.19} & \textbf{2.96} & \textbf{73.4} & \textbf{92.3} \\
    \bottomrule
    \end{tabular}
    }
    \vspace{1mm}
    \footnotesize{$\uparrow$ indicates higher variance than the full trajectory. ``—'' denotes undefined variance for symmetric tasks.}
\end{table}

\begin{figure}[tb]
    \centering
    \includegraphics[width=1.0\linewidth]{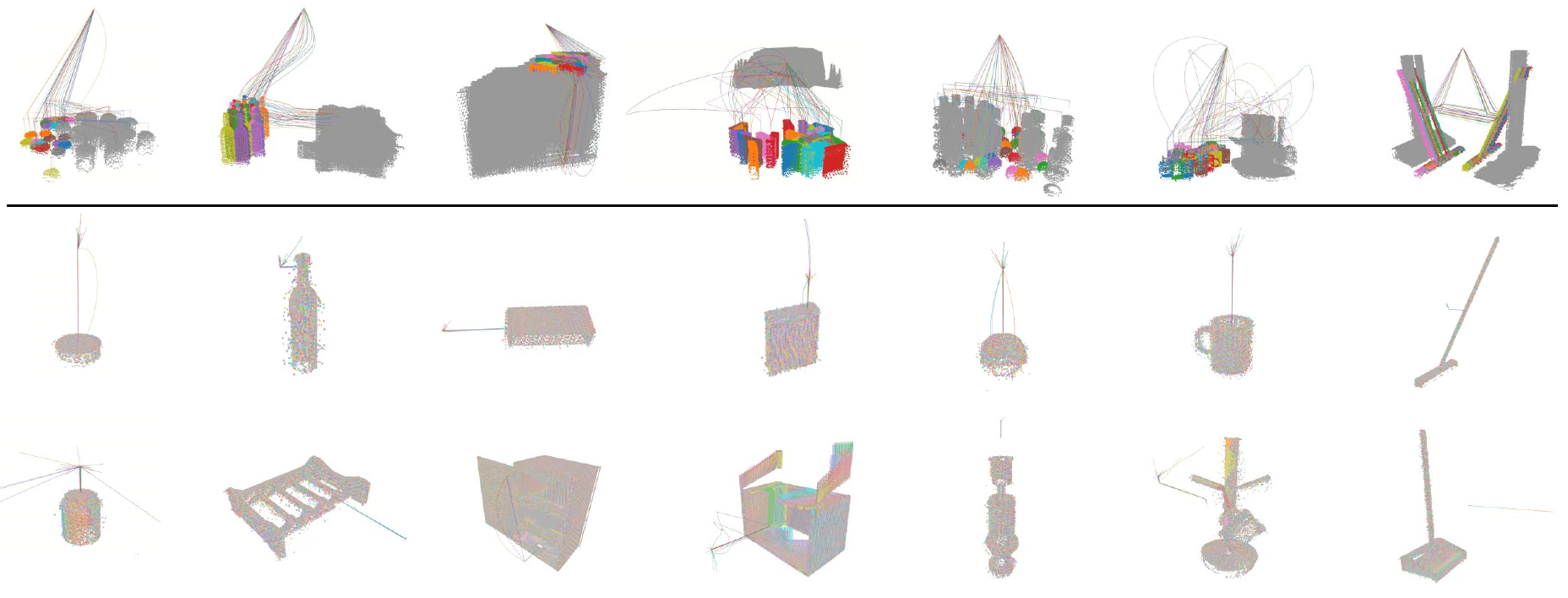}
    \caption{Visualization of interaction-centric formulation on 7 RLBench-18 tasks with 30 demonstrations each. Top: full end-effector trajectories in the world frame. Middle and bottom: extracted GOI and OOI interaction segments transformed into the corresponding source and target object frames. After canonicalization and temporal abstraction, interaction trajectories become substantially more consistent across demonstrations. Tasks from left to right: \textit{close\_jar}, \textit{stack\_wine}, \textit{put\_in\_safe}, \textit{put\_groceries}, \textit{screw\_bulb}, \textit{place\_cups}, \textit{sweep\_dust}.}
    \label{fig:app_vis_variance}
\end{figure}

To further illustrate the effect of interaction-centric formulation, we visualize demonstrations from 7 RLBench-18 tasks in Fig.~\ref{fig:app_vis_variance}. The first row shows full trajectories in the world frame, where demonstrations exhibit large spatial variability across object configurations. In contrast, after transforming trajectories into the corresponding GOI and OOI object frames, the extracted interaction segments become substantially more consistent across demonstrations. This visualization confirms that the proposed interaction segmentation isolates the geometrically constrained phases of manipulation while removing task-irrelevant transport variability.

\begin{figure}[tbp]
    \centering
    \includegraphics[width=1.0\linewidth]{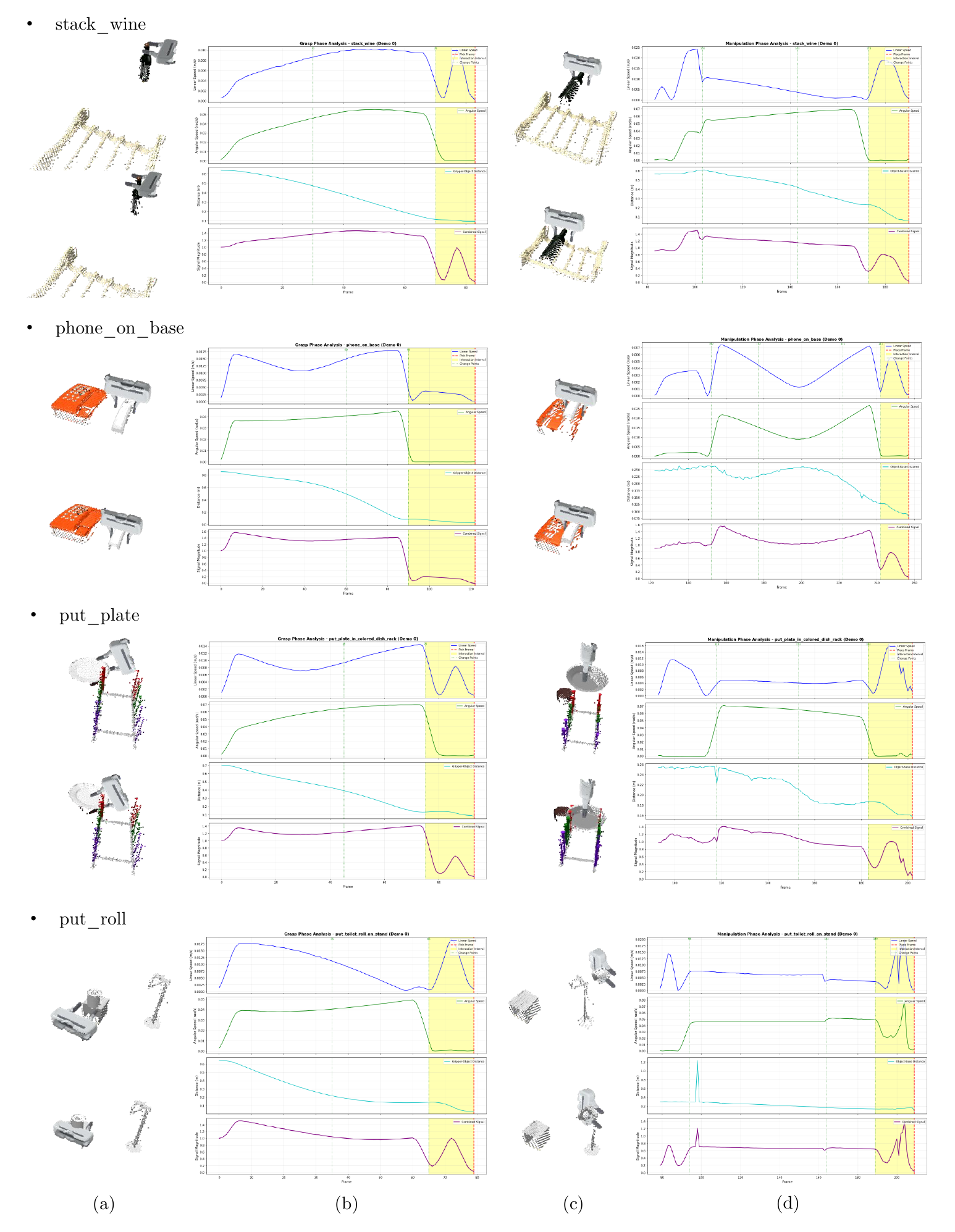}
    \caption{Visualization of interaction segment extraction of four RLBench tasks. For each task, we show the start and end frames of the interaction interval together with the corresponding signal evolution. (a) and (b) correspond to the GOI process. (c) and (d) correspond to the OOI process. (a) and (c): task-relevant point clouds at the start and end of the interaction interval. Top: start of the interaction segment; bottom: end of the interaction segment. (b) and (d): temporal evolution of segmentation signals, including gripper linear speed (blue), angular speed (green), inter-object distance (cyan), and the combined signal (purple). The shaded region indicates the detected interaction interval.}
    \label{fig:intervals}
\end{figure}

\begin{figure}[tbp]
    \centering
    \includegraphics[width=1.0\linewidth]{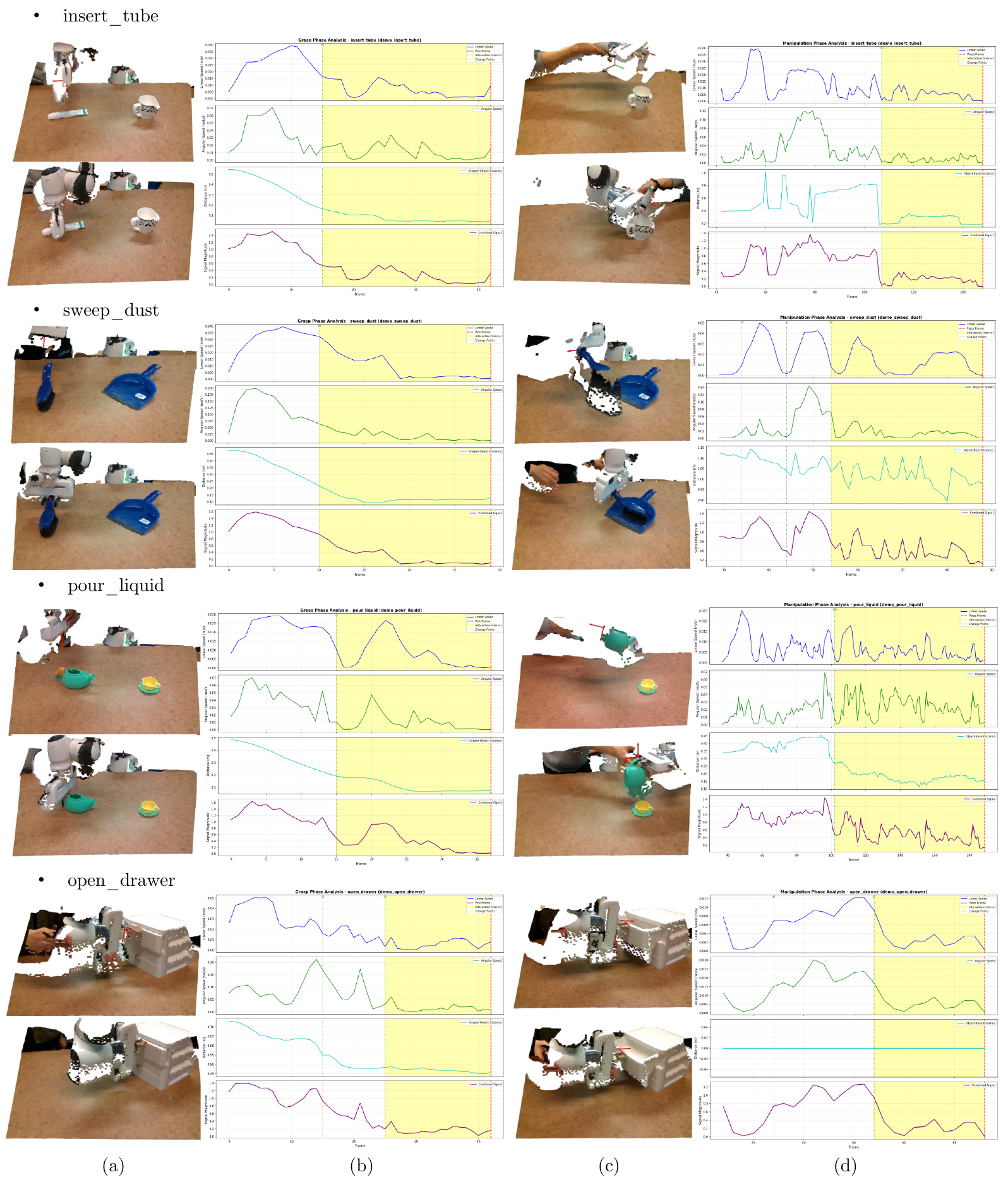}
    \caption{Visualization of interaction segment extraction of four real-world tasks. For each task, we show the start and end frames of the interaction interval together with the corresponding signal evolution. (a) and (b) correspond to the GOI process. (c) and (d) correspond to the OOI process. (a) and (c): task-relevant point clouds at the start and end of the interaction interval. Top: start of the interaction segment; bottom: end of the interaction segment. (b) and (d): temporal evolution of segmentation signals, including gripper linear speed (blue), angular speed (green), inter-object distance (cyan), and the combined signal (purple). The shaded region indicates the detected interaction interval.}
    \label{fig:intervals_real_world}
\end{figure}

To demonstrate that the extracted interaction segments align with intuitive manipulation phases, we visualize the extracted interaction segments for one demonstration from four representative RLBench tasks in Fig.~\ref{fig:intervals} and four real-world tasks in Fig.~\ref{fig:intervals_real_world}. The results show that the identified interaction intervals correspond well to semantically meaningful task phases. We further plot the evolution of the kinematic and spatial signals used for segmentation. As shown in Fig.~\ref{fig:intervals}, the proposed signals clearly capture the transition from unconstrained transportation to constrained object interaction, validating the effectiveness of our interaction segment identification method. In Fig.~\ref{fig:intervals_real_world}, although the real-world signals are noisier, the method remains sufficient to identify reasonable interaction intervals.

To demonstrate the stability of the interaction segment extraction method, we report the mean and standard deviation of interaction interval lengths across Relational Manipulation Benchmark tasks in Table~\ref{tab:intervals}. The relatively low variance within each task indicates that the extracted interaction segments are temporally consistent across demonstrations. At the same time, interval lengths vary across tasks in a semantically meaningful way. For example, tasks requiring precise alignment or sustained contact (e.g., \textit{screw\_nail}) exhibit longer interaction phases, while simpler interactions (e.g., \textit{meat\_on\_grill}) are shorter. This suggests that the proposed method captures task-dependent interaction structure rather than relying on fixed heuristics.

\begin{table}[ht]
    \centering
    \caption{Mean and standard deviation of interaction interval lengths across 10 RLBench tasks.}
    \label{tab:intervals}
    \begin{adjustbox}{max width=\textwidth}
    \begin{tabular}{l | c c}
    \toprule
    Tasks & GOI interval & OOI interval \\
    \midrule
    phone\_on\_base & 26.8 ± 9.0 & 15.6 ± 1.6 \\
    stack\_wine & 13.8 ± 1.5 & 19.8 ± 2.4 \\
    put\_plate & 15.6 ± 2.4 & 19.2 ± 1.0 \\
    put\_roll & 15.6 ± 2.4 & 19.2 ± 1.0 \\
    put\_rubbish & 17.8 ± 3.8 & 25.0 ± 5.7 \\
    put\_umbrella & 13.2 ± 2.1 & 24.2 ± 1.6 \\
    meat\_on\_grill & 16.4 ± 1.6 & 6.2 ± 3.5 \\
    put\_books & 12.2 ± 1.9 & 25.2 ± 3.7 \\
    screw\_nail & 15.6 ± 1.9 & 35.8 ± 2.6 \\
    turn\_tap & 29.8 ± 6.7 & 6.0 ± 3.1 \\
    \bottomrule
    \end{tabular}
    \end{adjustbox}
\end{table}

\subsection{Robustness to noisy pose estimation on RLBench} 
\label{sec: noise_test}
We evaluate the robustness of \textsc{Foci Policy} under noisy pose estimation on five representative RLBench tasks. All experiments are conducted in a single-view setting using the front camera. Although the reported results in Table~\ref{tab:simu_results} already include pose estimation errors from FoundationPose, we further perform a stress test by injecting additional noise to simulate more challenging perception conditions. Specifically, we perturb the estimated object poses with bounded uniform noise in both translation and rotation. Translation noise is sampled within $\pm \delta_t$ meters and rotation noise within $\pm \delta_r$ degrees along each axis, applied independently to each pose. We consider three noise levels: \textit{mild} ($\delta_t{=}0.5$cm, $\delta_r{=}2^\circ$), \textit{medium} ($\delta_t{=}1$cm, $\delta_r{=}5^\circ$), and \textit{hard} ($\delta_t{=}2$cm, $\delta_r{=}10^\circ$). We evaluate the final model checkpoints (trained on five demonstrations) over three random seeds, with 25 episodes per task. Results are reported in Table~\ref{tab:noise}.

\begin{table}[tb]
\centering
\caption{Success rates (\%) of \textsc{Foci Policy} on RLBench under increasing levels of pose estimation noise.}
\label{tab:noise}
\begin{adjustbox}{max width=\textwidth}
\begin{tabular}{l | ccccc | c}
    \toprule
    Noise Level & phone\_on\_base & stack\_wine & put\_umbrella & put\_roll & turn\_tap & Avg. \\
    \midrule
    None & \textbf{97.3 ± 1.9} & \textbf{100.0 ± 0.0} &  \textbf{52.0 ± 3.3} & \textbf{45.3 ± 5.0} & \textbf{100.0 ± 0.0} & \textbf{78.9} \\
    Mild & 96.0 ± 3.3 & 97.3 ± 1.9 & 44.0 ± 3.3 & 38.7 ± 5.0 & 96.0 ± 3.3 & 74.4 \\
    Medium & \textbf{97.3 ± 1.9} & \textbf{100.0 ± 0.0} & 36.0 ± 6.5 & 25.3 ± 7.5 & 88.0 ± 5.7 & 69.3 \\
    Hard & 90.7 ± 1.9 & 96.0 ± 3.3 & 24.0 ± 3.3 & 21.3 ± 5.0 & 86.7 ± 8.2 & 63.7 \\
    \bottomrule
\end{tabular}
\end{adjustbox}
\end{table}

The results show that \textsc{Foci Policy} exhibits graceful degradation under increasing pose noise, indicating a degree of robustness to imperfect perception. Tasks with looser geometric constraints, such as \textit{phone\_on\_base} and \textit{stack\_wine}, maintain high success rates even under hard noise, suggesting that coarse interaction patterns are sufficient for successful execution. In contrast, tasks requiring precise spatial alignment, such as \textit{put\_umbrella} and \textit{put\_roll}, show significant performance drops under severe perturbations (over 50\% relative decrease). These results suggest that while the interaction-centric formulation shows robustness to moderate noise, performance in precision-critical tasks still depends on reliable pose estimation. Future work may address this limitation by leveraging more advanced pose estimation methods, such as stereo-based approaches (e.g., FoundationStereo~\cite{wen2025stereo}), to improve geometric accuracy under challenging conditions.

\subsection{Additional Results on RLBench-18 Tasks}
\label{app:rlbench_18_details}
We provide detailed results on RLBench-18 in Table~\ref{tab:rlbench18_full}. The tasks are grouped by \textit{interaction segment grounding difficulty}: \textbf{Easy-to-ground} tasks involve geometrically distinctive objects with an identifiable contact event, enabling reliable interaction segment extraction and object canonicalization. In contrast, \textbf{hard-to-ground} tasks involve thin, symmetric, or heavily occluded objects that make reliable interaction grounding substantially more challenging.

To be more specific, tasks \textit{insert\_peg}, \textit{sort\_shape}, \textit{stack\_blocks}, \textit{screw\_bulb}, \textit{place\_cups}, and \textit{stack\_cups} involve small and symmetric objects that provide limited visual cues for reliable object canonicalization. Tasks \textit{open\_drawer}, \textit{put\_in\_drawer}, and \textit{put\_groceries} are heavily affected by occlusion under the single-camera setup.

Several observations can be made. First, despite using only a single demonstration, \textsc{Foci Policy} achieves strong performance on easy-to-ground tasks, substantially outperforming baselines in the low-data regime. Second, replacing estimated poses with ground-truth poses significantly improves performance on hard-to-ground tasks (7.8\% $\rightarrow$ 78.1\%), while performance on easy-to-ground tasks also improves from 77.9\% to 98.2\%. This large gap suggests that the primary challenge is not the interaction-centric formulation itself, but reliable interaction grounding under imperfect object canonicalization.

Overall, these results suggest that interaction-centric policies are highly data-efficient when task-relevant interactions can be reliably grounded, while robust interaction grounding remains a key bottleneck for general-purpose manipulation.


\begin{table}[tb]
\centering
\caption{
Detailed RLBench-18 results (1-demo setting). Tasks are grouped by 
interaction segment grounding difficulty: \textit{easy-to-ground} tasks have 
visually stable and geometrically localized interactions; 
\textit{hard-to-ground} tasks involve thin geometry, occlusion, or ambiguous interaction modes. 
``w/ GT Pose'' isolates formulation effectiveness from perception errors.
Success rate (\%) over 25 tests, averaged across 3 runs.
}
\label{tab:rlbench18_full}
\begin{adjustbox}{max width=\textwidth}
\begin{tabular}{l|ccccccccc|c}
\toprule
\multicolumn{11}{c}{\textit{Easy-to-Ground Interactions}} \\
\midrule
Method & close\_jar & meat\_grill & place\_wine & push\_buttons & turn\_tap
       & drag\_stick & slide\_color & sweep\_dust & put\_in\_safe & Avg. \\
\midrule
\textsc{Foci Policy}
  & 70.7\tiny{±1.9} & 100.0\tiny{±0.0} & 97.3\tiny{±1.9} & 96.0\tiny{±3.3}
  & 96.0\tiny{±0.0} & 73.3\tiny{±3.8} & 82.7\tiny{±1.9} & 44.0\tiny{±15.0}
  & 56.0\tiny{±3.3} & \textbf{79.6} \\
\quad(w/ GT Pose)
  & 88.0\tiny{±3.3} & 100.0\tiny{±0.0} & 100.0\tiny{±0.0} & 100.0\tiny{±0.0}
  & 98.7\tiny{±1.9} & 100.0\tiny{±0.0} & 97.3\tiny{±1.9} & 100.0\tiny{±0.0}
  & 100.0\tiny{±0.0} & \textbf{98.2} \\
Imagination Policy
  & 68.0\tiny{±0.0} & 50.7\tiny{±1.9} & 96.0\tiny{±0.0} & 82.7\tiny{±5.0}
  & 0.0 & 0.0 & 0.0 & 12.0\tiny{±5.7}
  & 9.3\tiny{±1.9} & 35.4 \\
3D Diffuser Actor
  & 0.0 & 0.0 & 1.3\tiny{±1.9} & 0.0
  & 16.0\tiny{±5.7} & 0.0 & 6.7\tiny{±9.4} & 16.0\tiny{±0.0}
  & 8.0\tiny{±5.7} & 5.3 \\
\midrule
\multicolumn{11}{c}{\textit{Hard-to-Ground Interactions}} \\
\midrule
Method & insert\_peg & open\_drawer & put\_in\_drawer & sort\_shape & screw\_bulb & put\_groceries & place\_cups & stack\_blocks & stack\_cups & Avg. \\
\midrule
\textsc{Foci Policy}
  & 0.0 & 0.0 & 0.0 & 6.7\tiny{±3.8}
  & 16.0\tiny{±3.3} & 13.3\tiny{±1.9} & 5.3\tiny{±1.9} & 29.3\tiny{±1.9}
  & 0.0 & \textbf{7.8} \\
\quad(w/ GT Pose)
  & 98.7\tiny{±1.9} & 94.7\tiny{±3.8} & 90.7\tiny{±5.0} & 29.3\tiny{±6.8}
  & 76.0\tiny{±0.0} & 60.0\tiny{±5.7} & 84.0\tiny{±11.3} & 88.0\tiny{±3.3}
  & 81.3\tiny{±1.9} & \textbf{78.1} \\
Imagination Policy
  & 1.3\tiny{±1.9} & 0.0 & 0.0 & 0.0
  & 2.7\tiny{±1.9} & 14.7\tiny{±6.8} & 5.3\tiny{±1.9} & 0.0
  & 0.0 & 2.7 \\
3D Diffuser Actor
  & 0.0 & 0.0 & 0.0 & 0.0
  & 0.0 & 0.0 & 0.0 & 0.0
  & 0.0 & 0.0 \\
\midrule
\multicolumn{9}{l|}{\textit{Overall Average}}
  & \textsc{Foci} & \textbf{43.7} \\
\multicolumn{9}{l|}{}
  & (w/ GT Pose)  & \textbf{88.2} \\
\multicolumn{9}{l|}{}
  & Imagination  & 19.0 \\
\multicolumn{9}{l|}{}
  & 3DDA     & 2.7  \\
\bottomrule
\end{tabular}
\end{adjustbox}
\end{table}

\subsection{Additional Results with One Demonstration and One Camera} 
We report simulation results on 10 RLBench tasks in Table~\ref{tab:one_demo_simu} under a highly constrained setting: one demonstration and a single front-view camera. Several key observations are: (1) \textsc{Foci Policy} achieves an average success rate of $63.0\%$, significantly outperforming all baselines while requiring less training time, demonstrating strong data efficiency. In contrast, action-centric methods such as 3D Diffuser Actor and RVT-2 largely fail, with near-zero success rates on most tasks, indicating poor generalization in low-data regimes. (2) Object-centric baselines (Imagination Policy, SPOT) perform better than action-centric methods on several tasks, confirming their advantage in data efficiency. However, their performance is inconsistent, suggesting sensitivity to demonstration quality. (3) For object-centric policies, compared to their multi-demonstration results in Table~\ref{tab:simu_results}, their performance on some tasks even improves with a single demonstration. This suggests that while object-centric representations are more data-efficient, they are also sensitive to demonstration quality and consistency across demonstrations. The variability across demonstrations suggests that learning from few demonstrations requires modeling multiple plausible interaction modes. This highlights the importance of multi-modal object-centric policies, as partially addressed by \textsc{Foci Policy}, and motivates further research in this direction. 

Overall, these results highlight the importance of object-centric interaction modeling for robust manipulation under extreme data constraints.

\begin{table}[tb]
\centering
\caption{Performance comparisons under one-demonstration and single-camera setting on RLBench tasks. Success rate (\%) over 25 evaluation episodes. Results are averaged over 3 runs. Training time is also reported for all methods, measured on a single NVIDIA RTX 2080 Ti GPU.}
\label{tab:one_demo_simu}
\begin{adjustbox}{max width=\textwidth}
\begin{tabular}{l | cccccccccc |cc}
    \toprule
    Method & phone\_on\_base & stack\_wine & put\_plate & put\_roll & put\_rubbish & put\_umbrella & meat\_on\_grill & put\_books & screw\_nail & turn\_tap & Avg. & Time. \\
    \midrule
    \textsc{Foci Policy} (Ours) & \textbf{96.0 ± 3.3} & \textbf{97.3 ± 1.9} & 14.7 ± 1.9 & \textbf{45.3 ± 5.0} & \textbf{46.7 ± 6.8} & \textbf{37.7 ± 4.8} & \textbf{98.7 ± 1.9} & \textbf{54.7 ± 8.2} & \textbf{42.7 ± 1.9} & \textbf{96.0 ± 0.0} & \textbf{63.0} & \textbf{0.45 h} \\
    Imagination Policy & 49.3 ± 3.8 & 96.0 ± 0.0 & \textbf{24.0 ± 0.0} & 5.3 ± 1.9 & 18.7 ± 1.9 & 16.0 ± 6.5 & 34.7 ± 5.0 & 2.7 ± 1.9 & 5.3 ± 1.9 & 0.0 & 25.2 & 3.67 h \\
    SPOT & 2.7 ± 3.8 & 22.7 ± 8.2 & 0.0 & 0.0 & 5.3 ± 1.9 & 0.0 & 81.3 ± 1.9 & 0.0 & 14.7 ± 5.0 & \textbf{96.0 ± 0.0} & 22.3 & 1.63 h \\
    3D Diffuser Actor & 0.0 & 1.3 ± 1.9 & 5.3 ± 1.9 & 0.0 & 0.0& 0.0 & 0.0 & 5.3 ± 1.9 & 0.0 & 16.0 ± 5.7 & 2.8 & 2.08 h \\
    RVT2 & 0.0  & 14.7 ± 10.0 & 1.3 ± 1.9 & 0.0 & 8.0 ± 0.0 & 0.0 & 4.0 ± 0.0 & 0.0 & 0.0 & 46.7 ± 8.2 & 7.5 & 3.20 h \\
    \bottomrule
\end{tabular}
\end{adjustbox}
\end{table}

\subsection{Additional Results on COLOSSEUM}
\label{app:results_colosseum}
To further evaluate the generalization capability of \textsc{Foci Policy}, we benchmark our method on COLOSSEUM~\cite{pumacay2024colosseum}, an extension of RLBench designed to assess robustness under diverse visual perturbations. COLOSSEUM introduces 12 categories of distribution shifts unseen during training, including variations in object color, texture, and size, as well as changes in lighting, background, distractors, and camera pose.

We evaluate \textsc{Foci Policy} on six representative tasks (Table~\ref{tab:colosseum_tasks}). Following our low-data setting, the policy is trained using only a single demonstration from the original (unperturbed) environment. For evaluation, each task--perturbation pair is tested over 25 trials and averaged across 3 runs.

Table~\ref{tab:colosseum_tasks} reports the success rates across tasks and perturbation types. We observe that \textsc{Foci Policy} maintains strong robustness under a wide range of visual variations, particularly for tasks with well-defined geometric interactions (e.g., \textit{meat\_on\_grill} and \textit{stack\_wine}). In contrast, performance degrades more significantly under large geometric shifts (e.g., object size changes), highlighting the sensitivity of interaction execution to geometric inconsistencies. Notably, despite being trained on a single demonstration, the policy demonstrates non-trivial generalization to unseen perturbations such as background changes, distractors, and camera pose variations. 

These results support our central hypothesis that focusing on interaction segments provides a more invariant and transferable representation for relational manipulation, enabling generalization beyond the training distribution without requiring large-scale data.

\begin{table}[tb]
\centering
\caption{Success rates (\%) of \textsc{Foci Policy} on COLOSSEUM under diverse visual perturbations. The policy is trained with a single demonstration on the original setting.}
\label{tab:colosseum_tasks}
\begin{adjustbox}{max width=\textwidth}
\begin{tabular}{l | cccccc}
    \toprule
    Task Name & basketball\_in\_hoop & close\_box & close\_laptop\_lid & meat\_on\_grill & put\_money\_in\_safe & stack\_wine \\
    \midrule
    No variations & 34.7 ± 3.8  & 36.0 ± 3.3 & 46.7 ± 8.2 & 100.0 ± 0.0 & 36.0 ± 6.5 & 100.0 ± 0.0 \\
    All Perturbations & 4.0 ± 0.0 & 6.7 ± 1.9 & 5.3 ± 1.9 & 100.0 ± 0.0 & 4.0 ± 0.0 & 32.0 ± 3.3 \\
    MO-COLOR & 20.0 ± 3.3 & 33.3 ± 6.8 & 41.3 ± 7.5 & 100.0 ± 0.0 & 33.3 ± 5.0 & 73.3 ± 1.9 \\
    RO-COLOR & 22.7 ± 8.2 & - & - & 100.0 ± 0.0 & 29.3 ± 11.5 & 100.0 ± 0.0 \\
    MO-TEXTURE & 22.7 ± 3.8 & - & - & - & 41.3 ± 6.8 & - \\
    RO-TEXTURE & - & - & - & - & 34.7 ± 1.9 & 96.0 ± 0.0 \\
    MO-SIZE & 8.0 ± 0.0 & 4.0 ± 0.0 & 8.0 ± 0.0 & 100.0 ± 0.0 & 0.0 & 90.7 ± 1.9 \\
    RO-SIZE & 17.3 ± 1.9 & - & - & - & - & 98.7 ± 1.9 \\
    Light Color & 17.3 ± 1.9 & 33.3 ± 3.8 & 41.3 ± 1.9 & 100.0 ± 0.0 & 29.3 ± 7.5 & 100.0 ± 0.0 \\
    Table Color & 18.7 ± 1.9 & 48.0 ± 3.3  & 36.0 ± 0.0 & 100.0 ± 0.0 & 24.0 ± 6.5 & 98.7 ± 1.9 \\
    Table Texture & 33.3 ± 1.9 & 25.3 ± 6.8 & 34.7 ± 0.0 & 100.0 ± 0.0 & 30.7 ± 1.9 & 100.0 ± 0.0 \\
    Distractor & 30.7 ± 3.8 & 30.7 ± 5.0 & 40.0 ± 6.5 & 100.0 ± 0.0 & 25.3 ± 10.0 & 97.3 ± 1.9 \\
    Background Texture & 40.0 ± 0.0 & 33.3 ± 5.0 & 34.7 ± 10.5 & 100.0 ± 0.0 & 30.7 ± 5.0 & 100.0 ± 0.0 \\
    Camera Pose & 26.7 ± 5.0 & 24.0 ± 6.5 & 37.3 ± 6.8 & 100.0 ± 0.0 & 25.3 ± 5.0 & 92.0 ± 0.0 \\
    \midrule
    Task Mean & 22.8 ± 3.5 & 27.5 ± 4.7 & 32.5 ± 5.7 & 100.0 ± 0.0 & 26.5 ± 6.5 & 90.7 ± 1.5 \\
    \bottomrule
\end{tabular}
\end{adjustbox}
\end{table}

\begin{table}[tb]
\centering
\caption{Comparison with state-of-the-art VLA models on COLOSSEUM under diverse visual perturbations. Our method achieves competitive performance using only a single demonstration.}
\label{tab:colosseum_comparation}
\begin{adjustbox}{max width=\textwidth}
\begin{tabular}{l | cccccccc }
    \toprule
    Method & No Variations & All Perturbations & MO-COLOR & RO-COLOR & MO-TEXTURE & RO-TEXTURE & MO-SIZE & RO-SIZE \\
    \midrule
    \textsc{Foci Policy} (Ours) & 58.9 ± 4.8 & 25.3 ± 1.7 & 50.2 ± 4.9 & 63.0 ± 7.1 & 32.0 ± 5.5 & 65.4 ± 1.3 & 35.1 ± 0.9 & 58.0 ± 1.9 \\
    BridgeVLA & 96.5 ± 2.5 & 20.1 ± 8.9 & 77.6 ± 2.9 & 87.0 ± 4.8 & 82.7 ± 6.3 & 90.7 ± 5.8 & 91.1 ± 6.4 & 79.4 ± 2.7 \\
    ActiveVLA & 96.9 ± 1.9 & 21.4 ± 5.9 & 79.0 ± 1.5 & 88.4 ± 3.7 & 84.5 ± 1.1 & 92.0 ± 0.7 & 92.0 ± 1.3 & 80.5 ± 2.5 \\
    \midrule
    Method & Light Color & Table Color & Table Texture & Distractor & Background Texture & Camera Pose & Data. & Avg. \\
    \midrule
    \textsc{Foci Policy} (Ours) & 53.5 ± 3.6 & 54.2 ± 3.2 & 54.0 ± 3.0 & 54.0 ± 5.6 & 56.5 ± 5.2 & 50.9 ± 4.8 & 1 demo & 50.8 \\
    BridgeVLA & 86.2 ± 6.4 & 94.2 ± 3.1 & 93.1 ± 2.7 & 79.4 ± 3.7 & 95.6 ± 3.3 & 96.0 ± 3.7 & 100 demo & 83.5 \\
    ActiveVLA & 86.8 ± 4.6 & 95.1 ± 2.4 & 94.1 ± 1.0 & 80.4 ± 2.2 & 96.4 ± 1.1 & 96.4 ± 3.6 & 100 demo & 84.6 \\
    \bottomrule
\end{tabular}
\end{adjustbox}
\end{table}

We further compare \textsc{Foci Policy} with state-of-the-art 3D Vision-Language-Action (VLA) models on COLOSSEUM, including BridgeVLA~\cite{li2025bridgevla} and ActiveVLA~\cite{liu2026activevla}. BridgeVLA maps multi-view observations to action predictions via a vision-language backbone, while ActiveVLA incorporates active perception to improve spatial reasoning. We report their results as provided in the original papers.

Results are summarized in Table~\ref{tab:colosseum_comparation}. As expected, VLA models trained with significantly more data (100 demonstrations per task) achieve higher overall success rates, outperforming our method in absolute performance. However, despite being trained with only a single demonstration (1\% of the data), \textsc{Foci Policy} achieves an average success rate of 50.8\%, corresponding to over 60\% of VLA performance. Notably, under the most challenging setting (\textit{All Perturbations}), our method outperforms both VLA models, demonstrating stronger robustness under severe distribution shifts.

These results further support our central claim that modeling interaction segments provides a more data-efficient and robust abstraction for relational manipulation, reducing reliance on large-scale demonstrations.

\subsection{Ablation Study}
\label{app:ablation}
\textbf{Interaction Segment Interval.} We evaluate the effectiveness of our automatic interaction segment extraction by comparing it with fixed interval lengths and a no-segmentation baseline. Specifically, we consider (i) fixed-length intervals (10, 15, 20 frames), and (ii) a \textit{None} setting that directly predicts key poses over the entire trajectory without segmentation. All experiments are conducted in a single-view setting using the front camera.

Results are shown in Table~\ref{tab:interval_ablation}. On relatively simple pick-and-place tasks (e.g., \textit{phone\_on\_base} and \textit{stack\_wine}), arbitrary segmentation or no segmentation can achieve reasonable performance ($>77\%$). However, for precision-sensitive tasks (e.g., \textit{put\_umbrella}, \textit{put\_roll}, and \textit{put\_books}), performance degrades significantly when using fixed intervals or no segmentation. In particular, the \textit{None} setting fails almost completely on \textit{put\_umbrella}, indicating that learning from full trajectories introduces substantial noise from task-irrelevant motions. Fixed intervals partially alleviate this issue but remain suboptimal due to misalignment with true interaction phases. In contrast, our automatic segmentation consistently achieves the best or near-best performance, demonstrating that accurately localizing interaction segments is critical for learning transferable manipulation behaviors.

\begin{table}[tb]
\centering
\caption{Success rates (\%) of \textsc{Foci Policy} on RLBench under different interaction segment definitions. \textit{Auto} denotes our interaction segment identification method; fixed intervals use predefined interval lengths; \textit{None} corresponds to learning from full trajectories without segmentation.}
\label{tab:interval_ablation}
\begin{adjustbox}{max width=\textwidth}
\begin{tabular}{l | ccccc | c}
    \toprule
    Interval Length & phone\_on\_base & stack\_wine & put\_umbrella & put\_roll & put\_books & Avg. \\
    \midrule
    Auto & \textbf{97.3 ± 1.9} & \textbf{100.0 ± 0.0} &  \textbf{52.0 ± 3.3} & 45.3 ± 5.0 & \textbf{53.3 ± 1.9} & \textbf{69.6} \\
    10 & 81.3 ± 3.8 & 96.0 ± 3.3 & 17.3 ± 1.9 & \textbf{46.7 ± 1.9} & 8.0 ± 0.0 & 49.9 \\
    15 & 96.0 ± 3.3 & 98.7 ± 1.9 & 42.7 ± 1.9 & \textbf{46.7 ± 3.8} & 41.3 ± 5.0 & 65.1 \\
    20 & 93.3 ± 5.0 & \textbf{100.0 ± 0.0} & 44.0 ± 3.3 & 38.7 ± 1.9 & 46.7 ± 1.9 & 64.5 \\
    None & 77.3 ± 5.0 & 89.3 ± 1.9 & 0.0 & 25.3 ± 7.5 & 1.3 ± 1.9 & 38.6 \\
    \bottomrule
\end{tabular}
\end{adjustbox}
\end{table}

\begin{figure}[tb]
    \centering
    \includegraphics[width=0.8\linewidth]{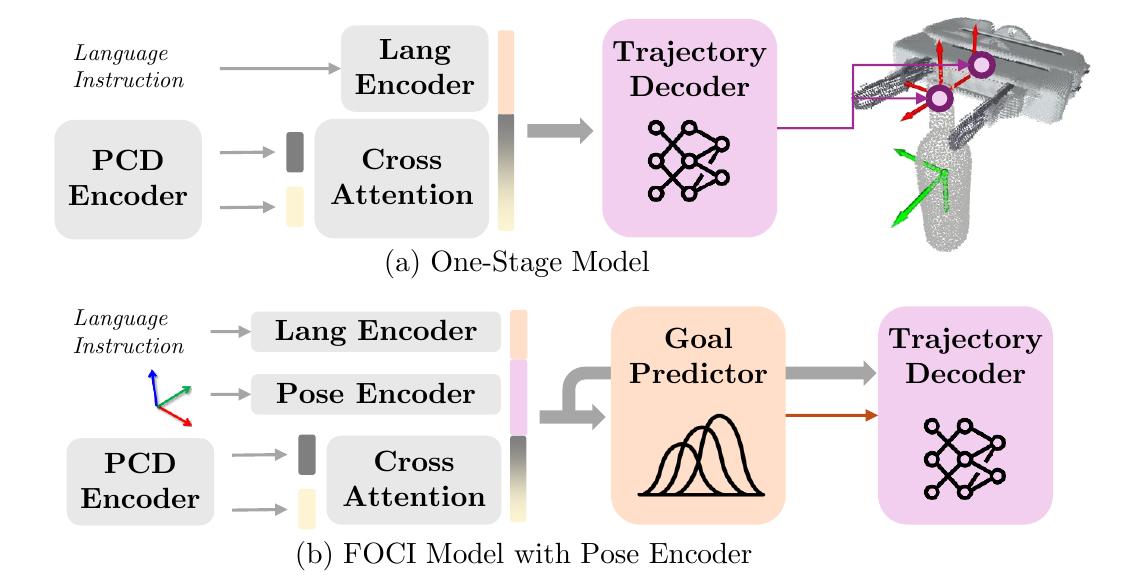}
    \caption{Model architecture ablations. (a) One-stage model without goal predictor. (b) Two-stage model augmented with pose features.}
    \label{fig:model_arch_ablation}
\end{figure}

\textbf{Model Architecture.} 
We first evaluate the effectiveness of the proposed two-stage formulation in Sec.~\ref{sec:model_architecture} by comparing it with a one-stage baseline. In the one-stage variant (Fig.~\ref{fig:model_arch_ablation}~(a)), the model directly predicts the full interaction segment $\tau_{\textnormal{rel}} = \{T^{(k)}_{\textnormal{rel}}\}_{k=1}^n$ conditioned on the fused representation $H$, without explicitly modeling the terminal state. In contrast, our two-stage model first predicts a distribution over the terminal interaction state and then generates the trajectory conditioned on the sampled goal. 

The one-stage model is trained with a trajectory regression objective:
\begin{equation}
    \mathcal{L} = \sum_{k=1}^{n}
    \left( \|\tilde{\mathbf{t}}_k - \mathbf{t}^{gt}_k\| + \lambda \|\tilde{\mathbf{r}}_k - \mathbf{r}^{gt}_k\| \right),
\end{equation}
where $\mathbf{t}$ and $\mathbf{r}$ denote translation and 6D rotation representations. 

We report results in Table~\ref{tab:arch_ablation} on five representative RLBench tasks. All experiments are conducted in a single-view setting using the front camera, with five demonstrations for training. Both models are trained for the same number of iterations. The two-stage model consistently outperforms the one-stage baseline across all tasks, with a larger gap on precision-sensitive tasks such as \textit{put\_roll}. This suggests that directly regressing the full trajectory can be difficult under limited data. By explicitly modeling the terminal interaction state, the two-stage formulation provides a clear supervision signal for goal alignment, which is particularly important for relational manipulation. We further evaluate the multi-modality of the two-stage model in Appendix~\ref{sec:generalizability}, where tasks share similar interaction patterns but differ in goal configurations.

\begin{table}[tb]
\centering
\caption{Ablation on model design. Success rates (\%) of \textsc{Foci Policy} under different architectural choices, including one-stage vs. two-stage formulation, the use of pose features, and random masking of point cloud tokens.}
\label{tab:arch_ablation}
\begin{adjustbox}{max width=\textwidth}
\begin{tabular}{l | ccccc | c}
    \toprule
    Model Variant & phone\_on\_base & stack\_wine & put\_umbrella & put\_roll & put\_books & Avg. \\
    \midrule
    Two-stage (w/o pose) & \textbf{97.3 ± 1.9} & \textbf{100.0 ± 0.0} &  52.0 ± 3.3 & \textbf{45.3 ± 5.0} & \textbf{53.3 ± 1.9} & \textbf{69.6} \\
    One-stage & 90.7 ± 5.0 & 90.7 ± 1.9 & 49.3 ± 10.0 & 16.0 ± 3.3 & 50.7 ± 5.0 & 59.5 \\
    Two-stage (w/ pose) & 93.3 ± 1.9 & 97.3 ± 3.8 & 53.3 ± 8.2 & 44.0 ± 3.3 & 45.3 ± 5.0 & 66.6 \\
    Two-stage (w/o masking) & 89.3 ± 3.8 & 96.0 ± 0.0 & \textbf{57.3 ± 5.0} & 26.7 ± 6.8 & 42.7 ± 5.0 & 62.4 \\
    \bottomrule
\end{tabular}
\end{adjustbox}
\end{table}

We further study the effect of incorporating pose features into the model (Fig.~\ref{fig:model_arch_ablation}~(b)). In addition to the geometric features $F^{\textnormal{src}}_{\textnormal{pcd}}$, $F^{\textnormal{tgt}}_{\textnormal{pcd}}$ and the language token $F_{\ell}$, we introduce a pose embedding $F^{\textnormal{src}}_{pose}$ for the source object. The pose is estimated from the demonstration using~\cite{wen2024foundationpose}, and is converted into a relative pose relative to the target object. After that, it is encoded using a two-layer MLP and projected to the same dimensionality as the point cloud features. The input tokens are then concatenated as
 \begin{equation}
    F = [F^{\textnormal{src}}_{\textnormal{pcd}}, F^{\textnormal{tgt}}_{\textnormal{pcd}}, F_{\ell}, F^{\textnormal{src}}_{pose}],
\end{equation}
which are processed by a stack of Transformer encoder layers with self-attention to obtain the fused representation $H$. The subsequent prediction follows the same two-stage formulation described in Sec.~\ref{sec:model_architecture}.

We report results in Table~\ref{tab:arch_ablation} on five representative RLBench tasks under the same setting as above. The model augmented with pose features achieves comparable performance to the version without pose input. This result suggests that explicit pose information provides limited additional benefit beyond point cloud geometry in our setting. We attribute this to the interaction-centric formulation: by focusing on interaction segments and modeling relative motion between objects, the policy primarily depends on local geometric relationships rather than the absolute object poses at the start of the task. As a result, incorporating pose features does not significantly improve performance, and we omit them in the final model for simplicity.

In Sec.~\ref{sec:model_architecture}, we apply random masking by dropping a proportion $\phi$ of point cloud cluster tokens during training. To evaluate its effect, we remove this masking strategy (denoted as ``w/o masking'' in Table~\ref{tab:arch_ablation}) and keep all other settings unchanged.

We report results in Table~\ref{tab:arch_ablation} on five representative RLBench tasks. Removing random masking leads to performance degradation on most tasks, particularly on \textit{put\_roll} and \textit{put\_books}, indicating reduced robustness to partial observations. We attribute this to the role of random masking as a form of regularization that encourages the model to rely on sparse and interaction-relevant local geometry. Without masking, the model becomes more sensitive to observation noise and missing regions at test time. An exception is observed on \textit{put\_umbrella}, where removing masking slightly improves performance. We hypothesize that for thin or elongated objects, aggressive token dropping may discard critical geometric details, limiting feature expressiveness. Overall, random masking improves robustness on most tasks, and we retain it in the final model.

\subsection{Special Case Study}

\begin{figure}[tb]
    \centering
    \includegraphics[width=1.0\linewidth]{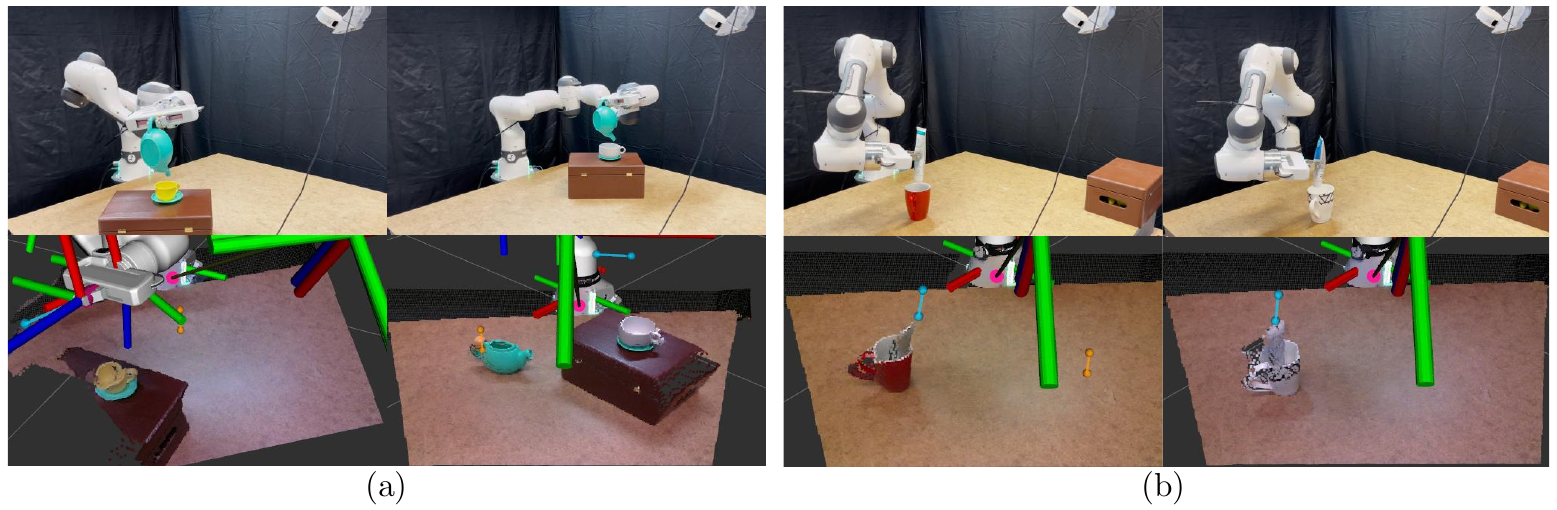}
    \caption{Generalization to unseen objects. \textsc{Foci Policy} transfers to novel object instances within the same category despite changes in appearance. Top row: snapshots at the place frame. Bottom row: observed point clouds with predicted gripper trajectories visualized in RViz2, where orange denotes GOI trajectories and cyan denotes OOI trajectories.(a) Generalization from a plastic cup to a visually distinct ceramic cup in the \textit{pour-liquid} task. (b) Robust insertion under variations in object color and appearance in the \textit{insert-tube} task.}
    \label{fig:generalizability}
\end{figure}

\textbf{Generalizability.}
\label{sec:generalizability} 
Leveraging category-level priors provided by pretrained pose estimators, \textsc{Foci Policy} generalizes to unseen objects with similar geometries and semantics. We illustrate two representative cases in Fig.~\ref{fig:generalizability}. In Fig.~\ref{fig:generalizability}(a), the policy is trained with a yellow plastic cup but successfully transfers to a visually distinct white ceramic cup in a zero-shot manner. In Fig.~\ref{fig:generalizability}(b), despite changes in color and appearance of both the toothpaste and the mug, the policy consistently performs the insertion task. These results highlight that our object-centric interaction representation captures task-relevant geometric relationships rather than instance-specific appearance, enabling robust generalization across object variations.

\begin{figure}[tb]
    \centering
    \includegraphics[width=1.0\linewidth]{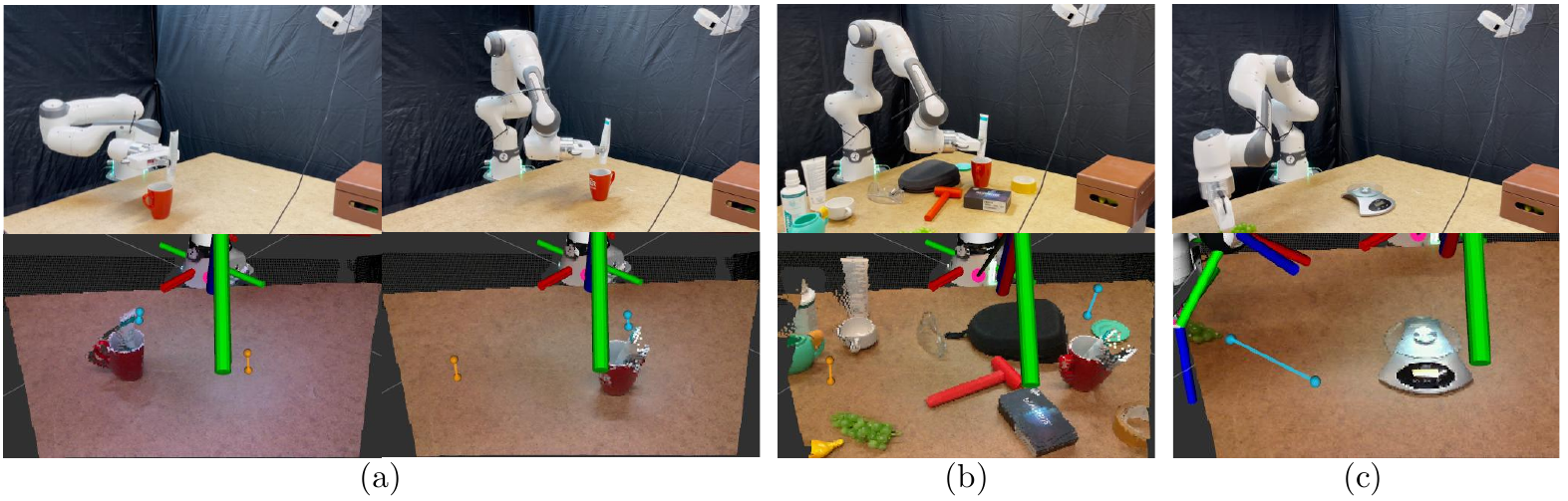}
    \caption{Robustness evaluation. (a) Multi-modal interaction generation under varying insertion configurations. (b) Robustness to cluttered scenes with distractor objects. (c) Robustness to partial observations, where the target object is partially outside the camera field of view.}
    \label{fig:robustness}
\end{figure}

\textbf{Robustness.} We demonstrate the robustness of \textsc{Foci Policy} under diverse challenging conditions in Fig.~\ref{fig:robustness}. In Fig.~\ref{fig:robustness}(a), we provide demonstrations of the \textit{insert-tube} task with varying insertion angles. Due to the multi-modal nature of the GMM-based goal predictor, the policy is able to generate different valid insertion trajectories conditioned on the relative configuration of the objects (i.e., varying mug positions and handle orientations), enabling successful task completion under diverse setups. In Fig.~\ref{fig:robustness}(b), we construct a cluttered scene with multiple distractor objects. Leveraging object segmentation, \textsc{Foci Policy} focuses on task-relevant objects and remains unaffected by irrelevant distractions, successfully completing the \textit{insert-tube} task. In Fig.~\ref{fig:robustness}(c), we consider partial observations where the target object (grape) is partially outside the field of view of the L515 camera. Benefiting from the generalization capability of the pretrained pose estimator, the system can still recover a reasonable object pose, allowing successful grasping despite incomplete observations.

\textbf{Obstacle Avoidance.} Benefiting from the interaction-centric formulation, \textsc{Foci Policy} naturally enables obstacle avoidance during manipulation, a capability that is difficult to achieve with end-to-end action-centric policies. Specifically, our framework follows a pipeline of motion planning $\rightarrow$ GOI inference $\rightarrow$ motion planning $\rightarrow$ OOI inference. The motion planner is responsible for generating collision-free trajectories that connect the predicted interaction poses. This design allows the system to directly leverage existing motion planning algorithms for obstacle avoidance without requiring additional learning.

\begin{figure}[tb]
    \centering
    \includegraphics[width=1.0\linewidth]{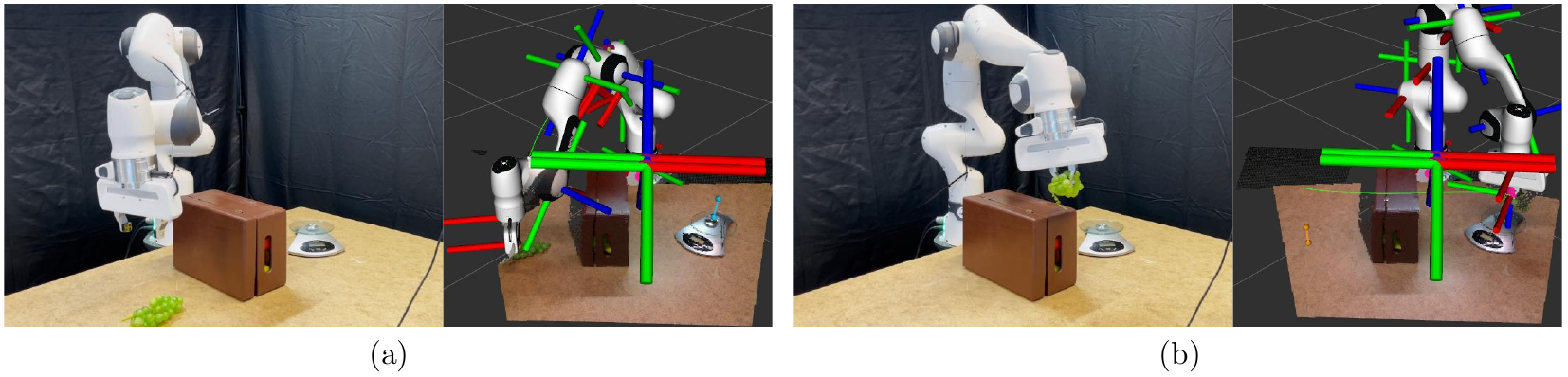}
    \caption{Obstacle avoidance in \textit{scale-grape} task. (a) GOI stage. (b) OOI stage. The green trajectory in RViz2 denotes the collision-free path generated by the motion planner, which connects the interaction poses predicted by \textsc{Foci Policy}.}
    \label{fig:obstacle_avoidance}
\end{figure}

We implement motion planning using cuRobo~\cite{sundaralingam2023curobo} in ROS2. As shown in Fig.~\ref{fig:obstacle_avoidance}, in the \textit{scale-grape} task, we place an obstacle between the grape and the scale. Although the model is trained with only a single demonstration and is never exposed to obstacle avoidance scenarios, it successfully completes the task in a zero-shot manner. This result highlights the effectiveness of combining segmented interaction prediction with classical motion planning.

\begin{figure}[tb]
    \centering
    \includegraphics[width=1.0\linewidth]{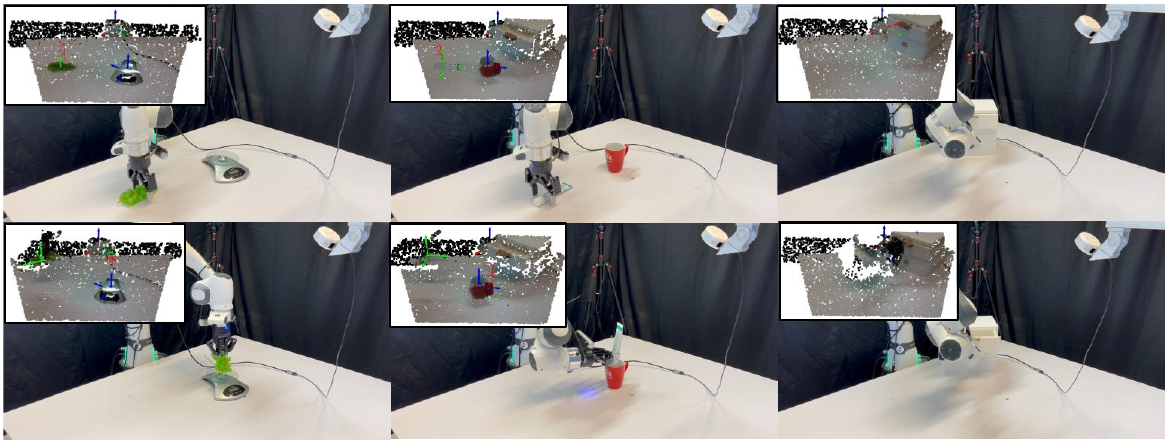}
    \caption{Cross-embodiment generalization. We transfer the policy from a Franka gripper to a Robotiq gripper with changes in scene appearance. Tasks from left to right: \textit{scale-grape}, \textit{insert-tube}, and \textit{open-drawer}. The first and second rows show the GOI and OOI frames, respectively. Each snapshot includes the corresponding point clouds and predicted results.}
    \label{fig:cross_embodiment}
\end{figure}

\textbf{Cross-Embodiment Generalization.}
Compared to action-centric policies, \textsc{Foci Policy} naturally supports cross-embodiment transfer. Since the policy predicts relative motion between task-relevant objects rather than robot-specific actions, the learned interaction trajectories can be reused across different robot embodiments. As shown in Fig.~\ref{fig:cross_embodiment}, we replace the original Franka gripper with a Robotiq gripper and modify the table appearance. Using only a single demonstration collected with the Franka gripper, we compute a fixed offset between the two grippers to adapt the predicted interaction segment. 

The results show that \textsc{Foci Policy} successfully transfers to the new embodiment and completes multiple tasks, including \textit{scale-grape}, \textit{insert-tube}, and \textit{open-drawer}, in a zero-shot manner. This demonstrates that the learned object-centric interaction representation is largely independent of robot embodiment and can serve as a reusable prior for different manipulation policies.

We further investigate whether \textsc{Foci Policy} can eliminate the need for demonstrations with robot data. To recover actions without access to robot states, we estimate hand pose trajectories directly from RGB-D observations using an off-the-shelf hand pose estimator~\cite{zhang2020mediapipe}. Following~\cite{papagiannis2025r}, we approximate the gripper pose by aligning a parallel-jaw gripper model (Robotiq) to key hand joints. Specifically, the gripper finger tips are aligned with the index and thumb tips, while its orientation is determined by the axis connecting these two points.

\begin{figure}[tb]
    \centering
    \includegraphics[width=0.8\linewidth]{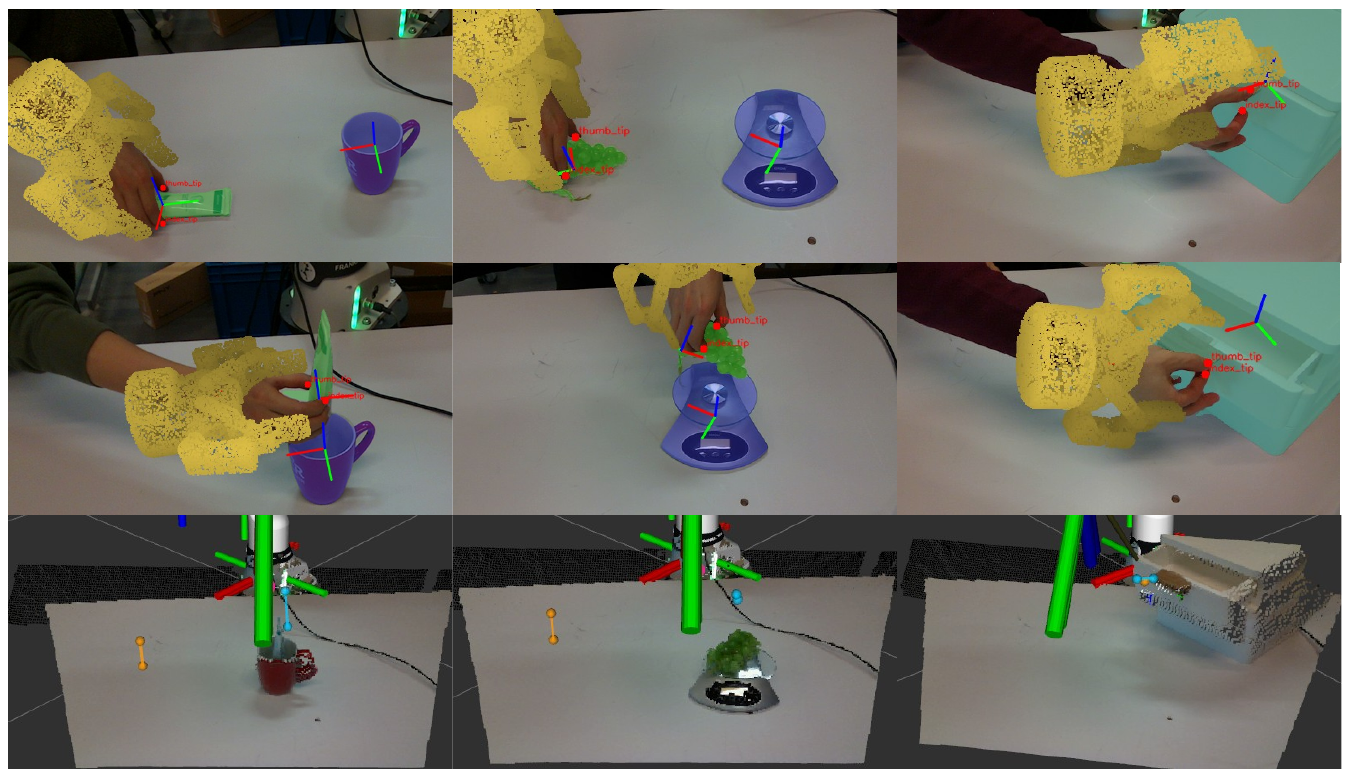}
    \caption{Robot-free demonstration pipeline and execution. Top two rows: processing of an RGB-D demonstration, including object segmentation, pose estimation, and hand pose extraction. The thumb and index fingertips are used to align and replace the gripper point cloud. Bottom row: predicted gripper trajectories during execution, where yellow indicates GOI prediction and cyan indicates OOI prediction.}
    \label{fig:cross_embodiment_experiment}
\end{figure}

We evaluate this pipeline on a Franka arm equipped with a Robotiq gripper across three tasks: \textit{scale-grape}, \textit{insert-tube}, and \textit{open-drawer}, using only a single human demonstration per task. As shown in Fig.~\ref{fig:cross_embodiment_experiment}, despite noise in hand pose estimation, the proposed interaction segment extraction and object-centric trajectory prediction effectively filter out high-frequency errors, resulting in stable manipulation behaviors during execution. These results suggest that \textsc{Foci Policy} can leverage weak, robot-free demonstrations by focusing on interaction structure rather than precise action trajectories.

\subsection{Failure Cases}
\label{sec:failure}
\begin{figure}[tb]
    \centering
    \includegraphics[width=1.0\linewidth]{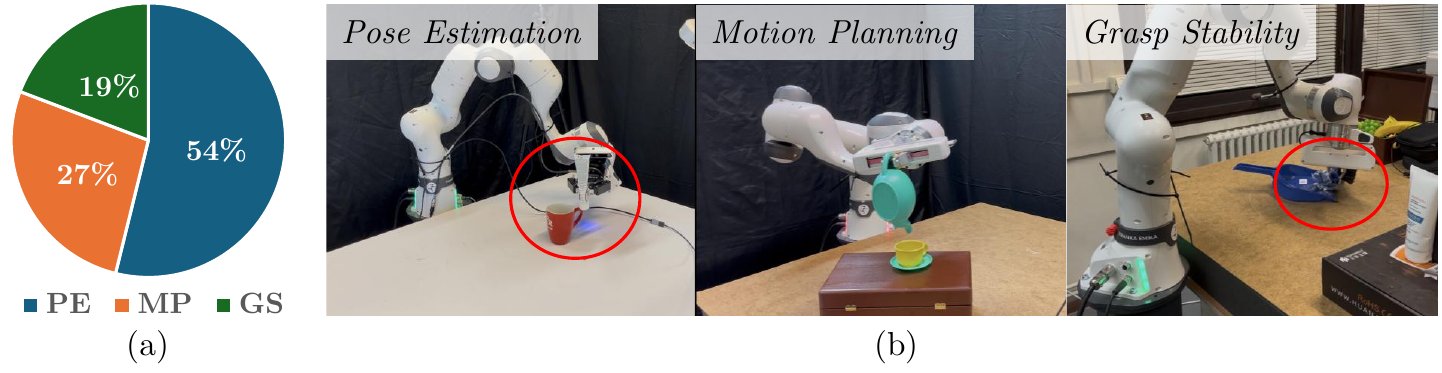}
    \caption{Failure analysis in real-world experiments. (a) Distribution of failure types across 26 trials. (b) Representative examples for each failure category: pose estimation error, motion planning failure, and grasp instability.}
    \label{fig:failure_count}
\end{figure}

We analyze failure cases from real-world experiments to better understand the limitations of our approach and identify directions for future improvement. As shown in Fig.~\ref{fig:failure_count}~(a), among 26 failure trials, 54\% are caused by pose estimation errors, 27\% by motion planning failures, and 19\% by grasp instability. Fig.~\ref{fig:failure_count}~(b) illustrates representative examples of each failure type. \textbf{Pose Estimation Errors:} errors in task-relevant object pose estimation can lead to incorrect interaction trajectories, resulting in failed grasps or inaccurate placement. \textbf{Motion Planning Failures:} in some cases, the robot fails to reach the predicted pose due to kinematic constraints, such as singularities or joint limits. \textbf{Grasp Instability:} during execution, the grasped object may shift or slip due to unmodeled dynamics, leading to failure in the subsequent manipulation stage.

Among these failure types, grasp instability can be mitigated by incorporating object tracking during execution, as our current implementation estimates object pose only once. Therefore, we focus our analysis on pose estimation and motion planning failures, which represent the primary bottlenecks in our system and provide more actionable insights for future improvements.

\begin{figure}[tb]
    \centering
    \includegraphics[width=1.0\linewidth]{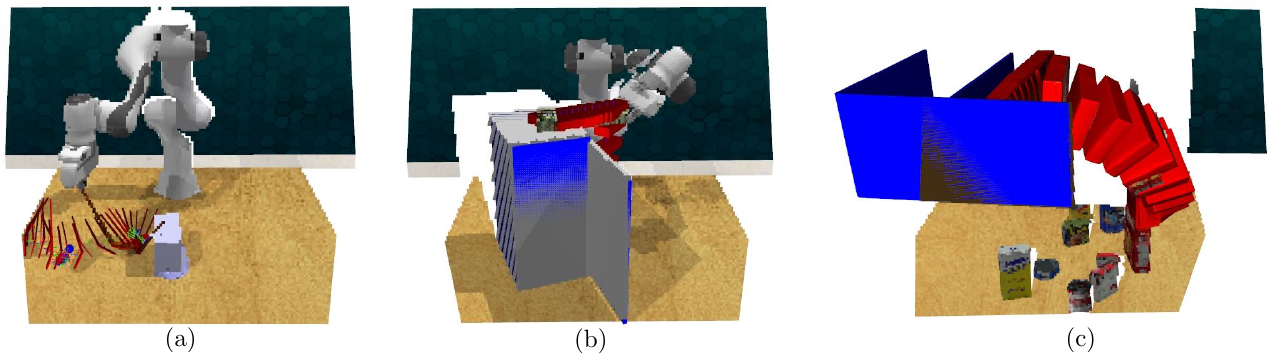}
    \caption{Illustration of pose estimation failure modes. (a) Small or thin objects with limited geometric features (task: \textit{Hockey}), where the slender object is difficult to localize reliably.  (b) Objects with symmetric or regular geometry (task: \textit{Put-Money-In-Safe}), leading to pose ambiguity.  (c) Heavily occluded or partially observed objects (task: \textit{Put-Groceries-In-Cupboard}), where the cupboard is partially visible from a single viewpoint, resulting in inaccurate pose estimation.}
    \label{fig:failure_fp}
\end{figure}

\textbf{Pose Estimation.} Our framework adopts FoundationPose~\cite{wen2024foundationpose} as the pose estimator due to its strong generalization to novel objects without task-specific fine-tuning. We observe several failure modes where pose estimation becomes unreliable: (a) small or thin objects with limited geometric features (e.g., the performance degradation observed in \textit{Insert-Peg} and \textit{Sort-Shape} in Table~\ref{tab:rlbench18_full}), (b) objects with symmetric or regular geometry leading to pose ambiguity, and (c) heavily occluded or partially observed objects. These cases are illustrated in Fig.~\ref{fig:failure_fp}. In such scenarios, pose errors directly propagate to the predicted interaction segments, resulting in misaligned trajectories and execution failures. Addressing these limitations may require more robust perception modules, such as point cloud completion~\cite{wei2025pcdreamerpointcloudcompletion} to handle partial observations, and semantic pose estimation~\cite{wang2023gsposecategorylevelobjectpose} to resolve geometric ambiguities.

\begin{figure}[tb]
    \centering
    \includegraphics[width=1.0\linewidth]{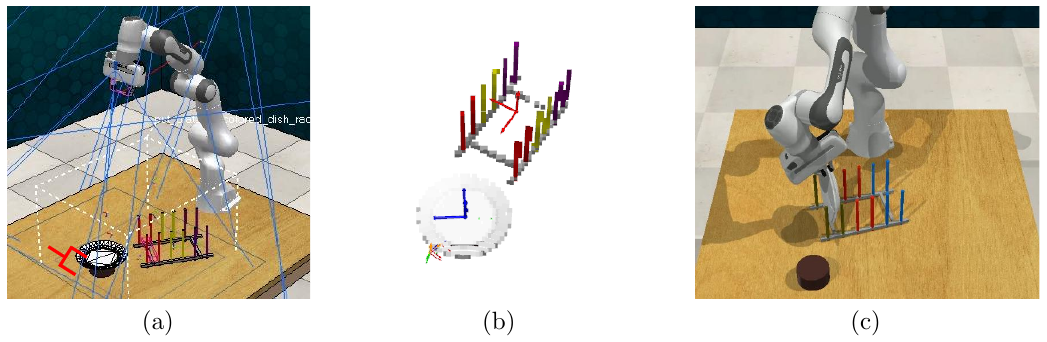}
    \caption{Failure cases in \textit{Put-Plate}. (a) and (b) Although the predicted interaction segment is semantically correct, it induces an unfavorable grasp configuration, preventing the motion planner from finding a feasible trajectory. (c) Generalization failure under task variations: the policy predicts an incorrect placement due to limited interaction diversity from a single demonstration.}
    \label{fig:failure_mp}
\end{figure}

\textbf{Motion Planning.} A representative failure case is \textit{Put-Plate}, as shown in Fig.~\ref{fig:failure_mp}(a) and (b). With only a single demonstration, the learned interaction is limited to a single mode. Although the predicted interaction segment is semantically correct, it may correspond to an unfavorable grasp configuration (e.g., grasping the plate outward the robot rather than toward), which prevents the motion planner from finding a feasible trajectory. Fig.~\ref{fig:failure_mp}(c) further illustrates another source of degradation. In RLBench, the valid placement slot in \textit{Put-Plate} varies across instances but is consistently indicated by color cues. Under a single-demonstration setting, the policy is exposed to only one interaction mode and thus cannot reliably disambiguate such variations, often generalizing to incorrect placement configurations. This limitation could be mitigated by incorporating multiple demonstrations or leveraging language to explicitly condition on different interaction modes.

These limitations are largely orthogonal to the proposed interaction-centric formulation, and instead reflect challenges in interaction coverage and conditioning under minimal data.


\section{Additional Method Details}
\subsection{Implementation Details}
\label{app:implementation_details}

\textbf{Data Preparation.} 
We construct training samples of the form $(P_{\textnormal{src}}, P_{\textnormal{tgt}}, \tau_{\textnormal{rel}}, \ell)$. 
For each demonstration in $\mathcal{D}$, task-relevant objects are segmented~\cite{ren2024grounded} and their point clouds are extracted from depth observations. Object poses $T_{\textnormal{src}}$ and $T_{\textnormal{tgt}}$ are estimated using~\cite{wen2024foundationpose}. To handle missing or noisy poses in single-view settings, we assume a rigid grasp and propagate object poses using gripper motion to fill gaps and reduce drift. We then convert to an object-centric representation by computing $T_{\textnormal{rel}} = T_{\textnormal{tgt}}^{-1} T_{\textnormal{src}}$, and expressing both point clouds in the target frame. Interaction segments $[t_\textnormal{s}, t_\textnormal{g}]$ are obtained using the method in Sec.~\ref{sec: interaction_segment}, from which $n$ relative waypoints $\tau_{\textnormal{rel}} = \{T^{(k)}_{\textnormal{rel}}\}_{k=1}^n$ are sampled. Each demonstration yields multiple training samples by pairing observations at different time steps with the same target trajectory. For multiple demonstrations, we optionally select the trajectory with the highest pose estimation confidence for supervision.

\textbf{Interaction Segment Identification.} 
For each demonstration, we construct kinematic and spatial signals to identify transitions from unconstrained motion to object interaction.  \textbf{Kinematic signals} include the gripper’s linear and angular speeds: $v(t) = \lVert d\mathbf{p}_g/dt \rVert$ and $\omega(t) = \lVert d\mathbf{q}_g/dt \rVert$, where $\mathbf{p}_g, \mathbf{q}_g$ denote the gripper position and orientation. \textbf{Spatial signals} capture object proximity via $d(t) = \lVert \mathbf{p}_{\textnormal{src}}(t) - \mathbf{p}_{\textnormal{tgt}}(t) \rVert$, where $\mathbf{p}_{\textnormal{src}}, \mathbf{p}_{\textnormal{tgt}}$ are object centroids. All signals are normalized and concatenated as:
\begin{equation}
    \tilde{S}(t) = [\tilde{v}(t), \tilde{\omega}(t), \tilde{d}(t)].
\end{equation}

Given $\tilde{S}(t)$, we identify the interaction interval $[t_\textnormal{s}, t_\textnormal{g}]$. The goal time $t_\textnormal{g}$ is obtained from gripper state changes or the end of the demonstration. To detect the interaction onset, we apply change-point detection using the PELT algorithm~\cite{truong2020selective} with an RBF kernel cost. It estimates a set of change points $\mathcal{T}=\{\tau_i\}_{i=1}^N$ (with $\tau_0=0$, $\tau_{N+1}=t_\textnormal{g}$) by minimizing:
\begin{equation}
    \min_{\mathcal{T}} \sum_{i=0}^{N} C\big(\tilde{S}(\tau_i:\tau_{i+1})\big) + \beta N,
\end{equation}
where $C(\cdot)$ measures within-segment variation under an RBF kernel, and $\beta$ controls the number of change points. 

For each subtask, the start time is defined as the last change point before $t_\textnormal{g}$:
\begin{equation}
    t_\textnormal{s} = \max\{\tau \in \mathcal{T} \mid \tau < t_\textnormal{g}\}.
\end{equation}

\textbf{Model Architecture.} 
As shown in Fig.~\ref{fig:model_arch}, the model consists of three components: masked geometric encoding, multi-modal fusion, and a two-stage trajectory predictor. Given $P_{\textnormal{src}}$ and $P_{\textnormal{tgt}}$, each point cloud is partitioned into $K_{\textnormal{src}}$ and $K_{\textnormal{tgt}}$ local clusters. A shared point-cloud encoder maps each cluster to a feature token. During training, a proportion $\phi$ of tokens is randomly dropped~\cite{pang2022masked} to improve robustness. The resulting tokens are $F^{\textnormal{src}}_{\textnormal{pcd}} = \{f^{\textnormal{src}}_i\}$ and $F^{\textnormal{tgt}}_{\textnormal{pcd}} = \{f^{\textnormal{tgt}}_j\}$. We use $10$ clusters with a masking ratio of $0.4$. Cross-attention is applied between the two token sets to capture interaction-relevant geometry. Language instruction is encoded as a token $F_{\ell}$ using a frozen CLIP~\cite{radford2021learning} text encoder. All tokens are concatenated as $F = [F^{\textnormal{src}}_{\textnormal{pcd}}, F^{\textnormal{tgt}}_{\textnormal{pcd}}, F_{\ell}]$ and processed by a Transformer encoder to obtain a fused representation $H$.

Conditioned on $H$, interaction prediction is factorized into two stages. A goal predictor models the terminal pose distribution using a Gaussian Mixture Model:
\begin{equation}
    p(T^{(n)}_{\textnormal{rel}} \mid H) = \sum_{m=1}^{M} \pi_m \mathcal{N}(T^{(n)}_{\textnormal{rel}} \mid \mu_m, \Sigma_m),
\end{equation}
where $\{\pi_m, \mu_m, \Sigma_m\}_{m=1}^M$ parameterize the GMM, and we use $M=5$ mixture components. A trajectory decoder then predicts intermediate poses conditioned on the sampled goal:
\begin{equation}
    p(\{T^{(k)}_{\textnormal{rel}}\}_{k=1}^{n-1} \mid H, T^{(n)}_{\textnormal{rel}}).
\end{equation}

The model is trained with supervision from ground-truth interaction segments. The goal predictor is optimized via negative log-likelihood:
\begin{equation}
    \mathcal{L}_{\text{goal}} = -\log \sum_{m=1}^{M} \pi_m \mathcal{N}(T^{(n)}_{\textnormal{rel}} \mid \mu_m, \Sigma_m),
\end{equation}
and the trajectory decoder minimizes:
\begin{equation}
    \mathcal{L}_{\text{traj}} = \sum_{k=1}^{n-1}
    \left( \|\tilde{\mathbf{t}}_k - \mathbf{t}^{gt}_k\| + \lambda \|\tilde{\mathbf{r}}_k - \mathbf{r}^{gt}_k\| \right),
\end{equation}
where $\mathbf{t}$ and $\mathbf{r}$ denote translation and 6D rotation. The final objective is:
\begin{equation}
    \mathcal{L} = \lambda_{\text{goal}} \mathcal{L}_{\text{goal}} + \lambda_{\text{traj}} \mathcal{L}_{\text{traj}}.
\end{equation}

\subsection{Tasks and Baseline Details}
\textbf{Relational Manipulation Benchmark Tasks.} We select eight challenging 3D tasks that require precise OOIs and geometric reasoning. The tasks cover a range of relational manipulation scenarios, including insertion (\textbf{\textit{Phone-on-Base}}, \textbf{\textit{Stack-Wine}}, \textbf{\textit{Put-Roll}}, \textbf{\textit{Put-Umbrella}}), placement under clutter or occlusion ( \textbf{\textit{Meat-on-Grill}}, \textbf{\textit{Put-Books}}), and articulated or multi-step manipulation (\textbf{\textit{Screw-Nail}}, \textbf{\textit{Turn-Tap}}). During the test, object poses are randomly sampled at the beginning of each episode, requiring policies to generalize to unseen object configurations.

\begin{itemize}
    \item \textbf{\textit{Phone-on-Base}:} grasp and insert a phone onto its base.
    \item \textbf{\textit{Stack-Wine}:} grasp a wine bottle and insert it into a rack slot.
    \item \textbf{\textit{Put-Roll}:} insert a toilet roll onto its stand with high precision.
    \item \textbf{\textit{Put-Umbrella}:} align an umbrella with its holder and insert it.
    \item \textbf{\textit{Meat-on-Grill}:} place a chicken leg onto a grill under occlusion.
    \item \textbf{\textit{Put-Books}:} retrieve a book from a stacked pile and place it onto a bookshelf.
    \item \textbf{\textit{Screw-Nail}:} align a screwdriver with a nail and perform rotational insertion.
    \item \textbf{\textit{Turn-Tap}:} rotate an articulated tap handle to a target orientation. 
\end{itemize}

\textbf{Simulation Baselines}. We compare \textsc{Foci Policy} against four strong multi-task baselines spanning object-centric and action-centric imitation learning paradigms:
\begin{itemize}
    \item \textbf{\textit{Imagination Policy}}~\cite{huang2024imagination} is a state-of-the-art object-centric method that generates interacting object point clouds and infers robot actions from the predicted goal configuration. It follows a similar pick–place decomposition but predicts only the final state.
    \item \textbf{\textit{SPOT}}~\cite{hsu2025spot} is an object-centric imitation learning method that employs a diffusion policy to generate object motion trajectories, from which robot actions are derived. 
    \item  \textbf{\textit{3D Diffuser Actor}}~\cite{ke20243d} extends Diffusion Policy by denoising action sequences conditioned on point cloud features.
    \item  \textbf{\textit{RVT-2}}~\cite{goyal2023rvt} is a vision-language policy that predicts key-frame gripper poses from multi-view visual features in a two-stage attention framework. 
\end{itemize}
Comparisons with the first two object-centric approaches evaluate the effectiveness and efficiency of directly predicting compact interaction pose sequences. The latter two baselines represent action-centric approaches that directly regress robot actions, enabling comparison between object-centric interaction modeling and action-centric policy learning under limited demonstration settings.

\textbf{RLBench-18 Tasks.} To further analyze the strengths and limitations of our method, we group the RLBench-18 tasks proposed by~\cite{shridhar2023perceiver} according to the \textit{interaction segment grounding difficulty}. The taxonomy reflects how reliably task-relevant OOIs can be extracted and canonicalized under a low-data and single-view setting, along two complementary dimensions:

\begin{itemize}
    \item \textbf{Perceptual grounding difficulty:} whether the object geometry and visibility support reliable object canonicalization and pose estimation.
    \item \textbf{Interaction structural ambiguity:} whether the interaction phase contains an identifiable contact event.
\end{itemize}

Based on these criteria, we divide the tasks into \textbf{easy-to-ground} and \textbf{hard-to-ground} categories.

\textbf{Easy-to-ground tasks.} These tasks involve geometrically distinctive objects, limited occlusion, and relatively well-defined interaction phases.

\begin{itemize}
    \item \textbf{\textit{Close-Jar}:} grasp the jar lid and rotate it to close the container.
    \item \textbf{\textit{Meat-off-Grill}:} pick a chicken leg from a grill under partial occlusion.
    \item \textbf{\textit{Place-Wine}:} grasp a wine bottle and insert it into a rack slot.
    \item \textbf{\textit{Push-Buttons}:} press the correct sequence of colored buttons.
    \item \textbf{\textit{Turn-Tap}:} rotate an articulated tap handle to a target orientation.
    \item \textbf{\textit{Drag-Stick}:} use a stick to drag a target object toward a goal region.
    \item \textbf{\textit{Slide-Color}:} slide a block to a colored target position.
    \item \textbf{\textit{Sweep-Dust}:} sweep scattered dust particles into a dustpan using a broom.
    \item \textbf{\textit{Put-in-Safe}:} place a dollar stack into a safe under partial occlusion.
\end{itemize}

\textbf{Hard-to-ground tasks.} These tasks involve thin, symmetric, or heavily occluded objects, making interaction grounding substantially more challenging.

\begin{itemize}
    \item \textbf{\textit{Insert-Peg}:} align and insert a thin peg into a colored pillar with high precision.
    \item \textbf{\textit{Sort-Shape}:} insert small geometric shapes into matching slots with high precision.
    \item \textbf{\textit{Screw-Bulb}:} align and screw a light bulb into its socket with high precision.
    \item \textbf{\textit{Put-Groceries}:} place a grocery item onto a shelf under heavy occlusion.
    \item \textbf{\textit{Place-Cups}:} insert a small cup onto a cup holder with high precision.
    \item \textbf{\textit{Open-Drawer}:} pull open a drawer under severe self-occlusion.
    \item \textbf{\textit{Put-in-Drawer}:} open a drawer and place a small cube into the drawer compartment.
    \item \textbf{\textit{Stack-Blocks}:} stack multiple small colored blocks onto a target block.
    \item \textbf{\textit{Stack-Cups}:} stack multiple symmetric cups with precise alignment.
\end{itemize}

\textbf{Real-World Tasks.}  We evaluate both methods on five relational manipulation tasks: 
\begin{itemize}
    \item \textbf{\textit{Scale-Grape}:} grasping a plastic grape and placing it onto a scale.
    \item \textbf{\textit{Sweep-Dust}:} grasping a broom to sweep crumpled paper into a dustpan.
    \item \textbf{\textit{Insert-Tube}:} inserting a toothpaste tube into a mug with precise alignment.
    \item \textbf{\textit{Pour-Liquid}:} pouring liquid from a teapot into a cup (for experimental convenience, ten small balls are placed inside the teapot, and success is defined as at least three landing in the cup).
    \item \textbf{\textit{Open-Drawer}:} opening a hinged drawer.
\end{itemize}

\textbf{Real-World Baselines.} To distinguish \textsc{Foci Policy} from pure trajectory replay, we implement a baseline termed \textit{Replay Policy}, inspired by~\cite{huang2025match}. This baseline extracts gripper poses together with $P_{\textnormal{src}}$ and $P_{\textnormal{tgt}}$ at four key states: pre-pick, pick, pre-place, and place. At test time, it applies Generalized ICP~\cite{segal2009icp} to estimate the rigid transformations of $P_{\textnormal{src}}$ and $P_{\textnormal{tgt}}$ between the demonstration and novel configurations, and transfers the demonstrated gripper poses accordingly to calculate robot actions. For a fair comparison, \textit{Replay Policy} uses the same interaction segment detection procedure to identify the four key states, and the number of interaction waypoints for our method is set to $n=2$. Both methods are provided with a single demonstration per task.

\subsection{Real-robot Experiment Details}
\label{app:real_world_details}
\begin{figure}[tb]
    \centering
    \includegraphics[width=1.0\linewidth]{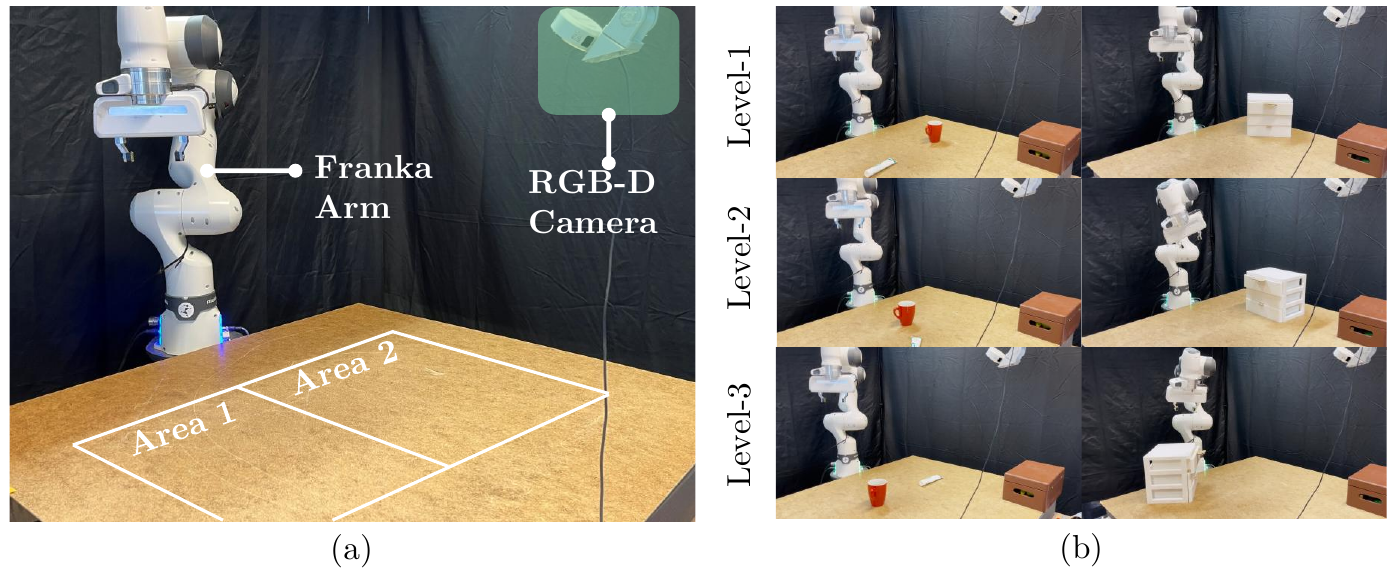}
    \caption{Real-robot experiment setup and evaluation protocol. (a) A single RGB-D camera provides a tabletop view for the Franka arm, with the workspace divided into GOI and OOI regions. (b) Example tasks (\textit{insert\_tube} and \textit{open\_drawer}) with three evaluation levels: Level-1 (small pose variations), Level-2 (larger pose variations), and Level-3 (swapped regions).}
    \label{fig:real_world_setting}
\end{figure}

As shown in Fig.~\ref{fig:real_world_setting}, experiments are conducted on a Franka Emika Panda robot equipped with a front-facing RealSense L515 RGB-D camera. The workspace consists of a $60 \,\text{cm} \times 50 \,\text{cm}$ tabletop, divided into separate GOI and OOI regions.

Each task is evaluated over 15 trials under three levels of variation. The first five trials (Level-1) use object poses with small perturbations from the demonstration. The next five trials (Level-2) introduce larger variations in object positions and orientations within their respective regions. The final five trials (Level-3) swap the GOI and OOI regions entirely, resulting in substantial changes in spatial configuration. This evaluation protocol progressively increases task difficulty and tests robustness to distribution shifts in object pose and workspace layout.

\begin{figure}[tb]
    \centering
    \includegraphics[width=0.9\linewidth]{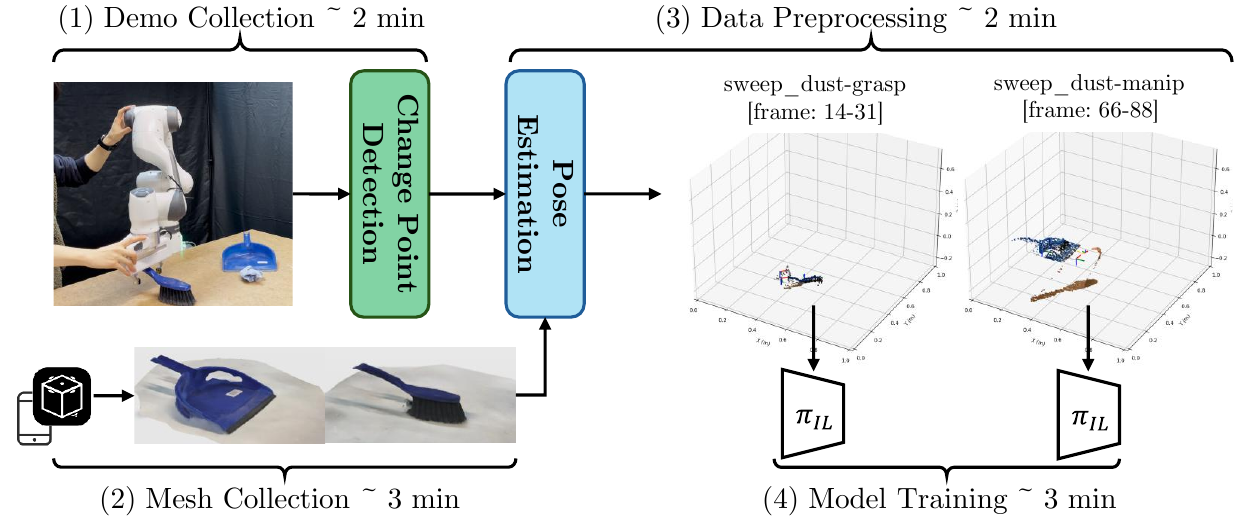}
    \caption{Real-world data collection and training pipeline. A new manipulation skill can be learned and deployed in under 10 minutes.}
    \label{fig:real_world_data}
\end{figure}

We illustrate the data collection and training pipeline, along with time consumption, in Fig.~\ref{fig:real_world_data}: (1) \textbf{Demonstration Collection.} We first enable impedance control on the Franka robot and kinesthetically guide it to perform the task. During this process, we record robot states, gripper states (open/close), and synchronized RGB-D observations from the RealSense L515 camera. This step takes approximately 2 minutes. (2) \textbf{Object Mesh Acquisition and Pose Tracking.} We use FoundationPose~\cite{wen2024foundationpose} for object pose estimation, which requires object meshes. Meshes are captured using a mobile phone with an AR scanning application, taking approximately 90 seconds per object. A lightweight post-processing step (scale verification and background removal in MeshLab) takes about 3 minutes. We observe that high-fidelity meshes are not required: as long as the scale and overall geometry are approximately correct, pose tracking remains reliable. (3) \textbf{Data Preprocessing.} The collected data is processed with change point detection and pose estimation, which takes approximately 2 minutes. (4) \textbf{Model Training.} Training the \textsc{Foci} model for a single task (including both GOI and OOI) takes about 3 minutes with a batch size of 8 on a single NVIDIA RTX 2080 Ti GPU. Overall, the entire pipeline (from demonstration to deployment) can be completed in under 10 minutes, enabling rapid skill acquisition in real-world settings.

\acknowledgments{This work was supported by Interne Fondsen KU Leuven/Internal Funds KU Leuven (C2E/24/034).}

\bibliography{example}  
    
\end{document}